%% file: main.tex
\documentclass[10pt,twocolumn,letterpaper]{article}

\usepackage[pagenumbers]{cvpr} 

\usepackage{graphicx}
\usepackage{amsmath}
\usepackage{amssymb}
\usepackage{booktabs}
\usepackage{multirow}
\usepackage{enumitem}

\usepackage[accsupp]{axessibility}  
\usepackage{xurl}

\newcommand{\cut}[1]{}

\newcommand{\circled}[1]{\raisebox{.5pt}{\textcircled{\raisebox{-.9pt} {{\small #1}}}}}

\AtBeginDocument{
  \abovedisplayskip     =0.5\abovedisplayskip
  \abovedisplayshortskip=0.5\abovedisplayshortskip
  \belowdisplayskip     =0.5\belowdisplayskip
  \belowdisplayshortskip=0.5\belowdisplayshortskip}

\iffalse

\newcommand{\nsection}[1]{\section{#1}}
\newcommand{\nsubsection}[1]{\subsection{#1}}

\else

\newcommand{\nsection}[1]{\vspace{-0.2cm}\section{#1}\vspace{-0.2cm}}
\newcommand{\nsubsection}[1]{\vspace{-0.05cm}\subsection{#1}\vspace{-0.1cm}}

\fi

\usepackage[pagebackref,breaklinks,colorlinks]{hyperref}

\usepackage[capitalize]{cleveref}
\crefname{section}{Sec.}{Secs.}
\Crefname{section}{Section}{Sections}
\Crefname{table}{Table}{Tables}
\crefname{table}{Tab.}{Tabs.}

\def\cvprPaperID{10899} 
\def\confName{CVPR}
\def\confYear{2022}

\newcommand{\newpart}[1]{#1}

\makeatletter
\newcommand\notsotiny{\@setfontsize\notsotiny{6}{7}}
\makeatother
\begin{document}

\title{Towards Driving-Oriented Metric for Lane Detection Models}

\author{Takami Sato\\
University of California, Irvine\\
{\tt\small takamis@uci.edu}
\and
Qi Alfred Chen\\
University of California, Irvine\\
{\tt\small alfchen@uci.edu}
}
\maketitle


\input{src/abstract}



\input{src/introduction}

\input{src/background}
\input{src/methodology}
\input{src/experiments}

\input{src/discussion}

\input{src/conclusion}

\vspace{-0.3cm}
\section*{Acknowledgments}
\vspace{-0.2cm}
\newpart{
This research was supported in part by the NSF CNS-1850533, CNS-1932464, CNS-1929771, CNS-2145493, and USDOT UTC Grant 69A3552047138.}
\vspace{-0.1in}

{\small
\bibliographystyle{ieee_fullname}
\bibliography{main.bib}
}

\appendix
\input{src/appendix}

\end{document}

%% file: src/abstract.tex
\begin{abstract}
After the 2017 TuSimple Lane Detection Challenge, its dataset and evaluation based on accuracy and F1 score have become the de facto standard to measure the performance of lane detection methods. 
While they have played a major role in improving the performance of lane detection methods, the validity of this evaluation method in downstream tasks has not been adequately researched.
In this study, we design 2 new driving-oriented metrics for lane detection: End-to-End Lateral Deviation metric (E2E-LD) is directly formulated based on the requirements of autonomous driving, a core downstream task of lane detection; Per-frame Simulated Lateral Deviation metric (PSLD) is a lightweight surrogate metric of E2E-LD.
To evaluate the validity of the metrics, we conduct a large-scale empirical study with 4 major types of lane detection approaches on the TuSimple dataset and our newly constructed dataset Comma2k19-LD.
Our results show that the conventional metrics have strongly negative correlations ($\leq$-0.55) with E2E-LD, meaning that some recent improvements purely targeting the conventional metrics may not have led to meaningful improvements in autonomous driving, but rather may actually have made it worse by overfitting to the conventional metrics. 
As autonomous driving is a security/safety-critical system, the underestimation of robustness hinders the sound development of practical lane detection models. We hope that our study will help the community achieve more downstream task-aware evaluations for lane detection.
\end{abstract}

%% file: src/introduction.tex
\vspace{-0.15in}
\nsection{Introduction} \label{sec:introduction}

Lane detection is one of the key technologies today for realizing autonomous driving.
For lane detection, camera is the most frequently used sensor because it is a natural choice as lane lines are visual patterns~\cite{hillel2014recent}. Like most other computer vision areas, lane detection has been significantly benefited from the recent advances of deep neural networks (DNNs). 
In the 2017 TuSimple Lane Detection Challenge~\cite{tusimple}, DNN-based lane detection shows substantial performance as all top 3 teams opt for DNN-based lane detection. After this competition, its dataset and evaluation method based on accuracy and F1 score became the \textit{de facto} standard in lane detection evaluation. These metrics are inherited by the subsequent datasets~\cite{pan2018spatial, llamas2019}.

\begin{figure}[t!]
\centering
\includegraphics[width=\linewidth]{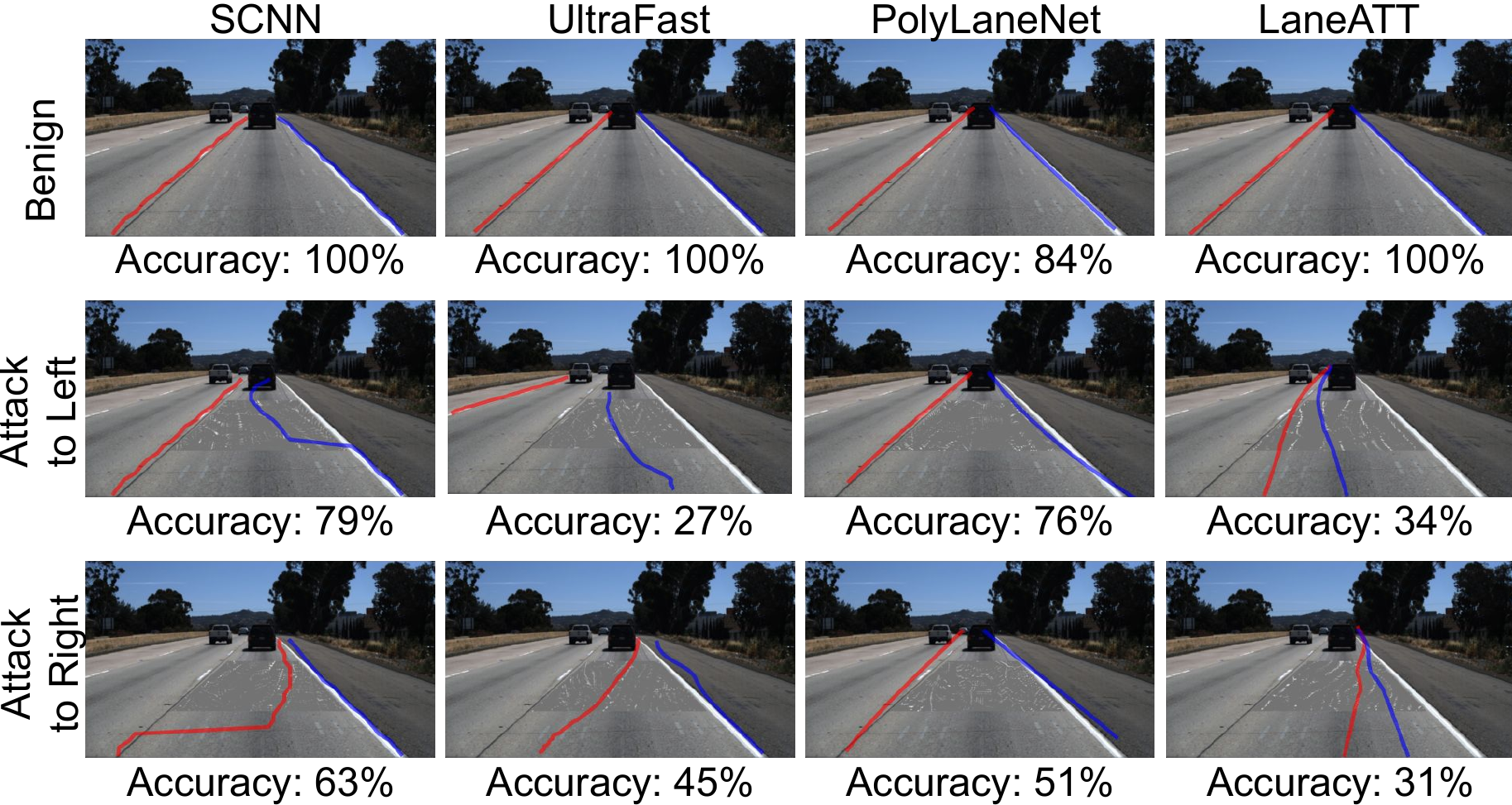}
\caption{Examples of lane detection results and the accuracy metric in benign and adversarial attack scenarios on TuSimple Challenge dataset~\cite{tusimple}. As shown, the conventional accuracy metric does not necessarily indicate drivability if used in autonomous driving, the core downsteam task. For example, SCNN always has higher accuracy than PolyLaneNet, but its detection results are making it much harder to achieve lane centering (detailed in~\S\ref{sec:end-to-end}).
}
\label{fig:poc_tusimple}
\label{fig:poc_tusimple_top}
\vspace{-0.2in}
\end{figure}

However, the validity of this evaluation method in practical contexts, i.e., whether this is representative of practicality in real-world downstream applications, has not been adequately researched. 
Specifically, the main real-world applications of lane detection are for autonomous driving (AD), e.g., online detection for automated lane centering (for lower-level AD such as in Telsa AutoPilot~\cite{tesla2020autopilot}), and offline detection for high-definition map creation (for both low-level~\cite{gmsupercruise} and high-level AD~\cite{TANG2021107623}). With such an application domain as its main target, the robustness of lane detection is highly critical as errors from it could be fatal. 
Unfortunately, we find that the conventional evaluation metrics (i.e., accuracy and F1 score) have limitations to correctly reflect the performance of lane detection models in such main downstream application domain, especially in more challenging scenarios (e.g., when under adversarial attacks). Fig.~\ref{fig:poc_tusimple_top} shows a few such examples that motivate this study. In the adversarial attack settings, the lane lines detected by SCNN~\cite{pan2018spatial} are largely disrupted, but their performance measured by the conventional accuracy metric is always higher than the one of PolyLaneNet~\cite{tabelini2021polylanenet}, which are generally aligned with actual lane lines (and indeed lead to less lane center deviation than SCNN when used with driving models as quantified later in~\S\ref{sec:end-to-end}). \newpart{In the benign settings, PolyLaneNet has the lowest accuracy and is underestimated, despite its seemingly perfect detection for humans.}
As lane detection has been evaluated using mainly relatively clean and homogeneous driver's view images, it is not easy to identify such a great discrepancy at the metric level. Considering the criticality of robust lane detection to correct and safe AD, it is important to address such a metric-level limitation since (1) the cornerstone of real-world deployment and commercialization of AD today is exactly on the handling of those more challenging driving scenarios~\cite{yu2020bdd100k, dosovitskiy2017carla, jain2021autonomy}; and (2) with increasingly more discoveries of physical-world adversarial attack on lane detection in AD context~\cite{sato2020hold, jing2021tencent}, it is desired to have a more downstream task-aware performance metric when judging the model robustness (and its enhancement).

Motivated by such critical needs, we design 2 new driving-oriented metrics, 
\textit{End-to-End Lateral Deviation metric} (E2E-LD) and \textit{Per-frame Simulated Lateral Deviation metric} (PSLD), to measure the performance of lane detection models in AD, especially in Automated Lane Centring (ALC), a Level-2 driving automation that automatically steers a vehicle to keep it centered in the traffic lane~\cite{sae2018}. E2E-LD is designed directly based on the requirements of driving automation by ALC.
PSLD is a lightweight surrogate metric of E2E-LD that estimates the impact of lane detection results on driving from a single frame. This per-frame lightweight design allows the metric to be usable during upstream lane detection model training.
To evaluate the validity of the metrics, we conduct a large-scale empirical study of the 4 major types of lane detection approaches on the TuSimple dataset and our newly constructed dataset, Comma2k19-LD, which contains both lane line annotation and driving information. To simulate corner-case but physically-realizable scenarios as in Fig.~\ref{fig:poc_tusimple} for lane detection, we utilize and extend physical-world adversarial attacks on ALC~\cite{sato2020hold}. We formulate attack objective functions to fairly generate adversarial attacks against the 4 major types of lane detection approaches.
Throughout this study, we find that the conventional metrics have strongly negative correlations ($r \leq$-0.55) with E2E-LD in the benign scenarios, meaning that some recent improvements purely targeting the conventional metrics may not have led to meaningful improvements in AD, but rather may actually have made it worse by overfitting to the conventional metrics.
In the attack scenarios, while we observe a slight positive correlation ($r \leq$0.08), it is not statistically significant. Consequently, we find that the conventional metrics tend to overestimate less robust models. On the contrary, our newly-designed PSLD metric is always strongly positively correlated with E2E-LD ($r \geq$0.38), and all correlations are statistically significant ($p \leq$ 0.001). 

While the TuSimple Challenge dataset and its evaluation metrics have played a substantial role in developing performant lane detection methods, the recent improvement on the conventional metrics does not lead to the improvement on the core downstream task AD. We thus want to inform the community of such limitations of the conventional evaluation and facilitate research to conduct more downstream task-aware evaluation for lane detection, as the gap between upstream evaluation metrics and downstream application performance may hinder the sound development of lane detection methods for real-world application scenarios. 

In summary, our contributions are as follows:
\vspace{-\topsep}
\begin{itemize}[leftmargin=0.2in]
\setlength{\itemsep}{0pt}
\setlength{\parskip}{0pt}
\item  We design 2 new driving-oriented metrics, E2E-LD and PSLD, that can more effectively measure the performance of lane detection models when used for AD, their core downstream task.
\item We design a methodology to fairly generate physical-world adversarial attacks against the 4 major types of lane detection models.
\item We build a new dataset Comma2k19-LD that contains lane annotations and driving information.
\item We are the first to conduct a large-scale empirical study to measure the capability of 4 major types of lane detection models in supporting AD. 
\item We highlight and discuss the critical limitations of the conventional evaluation and demonstrate the validity of our new downstream task-aware metrics.
\vspace{-\topsep}
\end{itemize}
\textbf{Code and data release.} All our codes and datasets are available in our project websites~\footnote{
\footnotesize{
\url{https://github.com/ASGuard-UCI/ld-metric}\\
\url{https://sites.google.com/view/cav-sec/ld-metric}}
}.

%% file: src/background.tex
\nsection{Related Work}
\vspace{0.05in}
\nsubsection{DNN-based Lane Detection} \label{sec:lane_detection}
We taxonomize state-of-the-art DNN-based lane detection methods into 4 approaches. Similar taxonomy is also adopted in prior works~\cite{tabelini2021cvpr, liu2021CondLaneNet}.

\textbf{Segmentation approach.} Segmentation approach handles lane detection as a segmentation task, which classifies whether each pixel is on a lane line or not. Since this approach achieved the state-of-the-art performance in the 2017 TuSimple Lane Detection Challenge~\cite{tusimple} (all top-3 winners adopt the segmentation approach~\cite{pan2018spatial, hsu2018learning, neven2018towards}), it has been applied in many recent lane detection methods~\cite{zheng2020resa, hou2019learning, zheng2021RESA}. This segmentation approach is also used in the industry. A reverse-engineering study reveals that Tesla Model S adopts this segmentation-based approach~\cite{jing2021tencent}.
The major drawback of this approach is its higher computational and memory cost than the other approaches. Due to the nature of the segmentation approach, it needs to predict the classification results for every pixel, the majority of which is just background. Additionally, this approach requires a postprocessing step to extract the lane line curves from the pixel-wise classification result.

\textbf{Row-wise classification approach.} This approach~\cite{qin2020ultra, yoo2020end, hou2020inter, liu2021CondLaneNet} leverages the domain-specific knowledge that the lane lines should locate the longitudinal direction of driving vehicles and should not be so curved to have more than 2 intersections in each row of the input image. Based on the assumption, this approach formulates the lane detection task as multiple row-wise classification tasks, i.e., only one pixel per row should have a lane line. Although it still needs to output classification results for every pixel similar to the segmentation approach, this divide-and-conquer strategy enables to reduce the model size and computation while keeping high accuracy. For example, UltraFast\cite{qin2020ultra} reports that their method can work at more than 300 FPS with a comparable accuracy 95.87\% on the TuSimple Challenge dataset~\cite{tusimple}. On the other hand, SAD~\cite{hou2019learning}, a segmentation approach, works at 75 frames per second with 96.64\% accuracy.
This approach also requires a postprocessing step to extract the lane lines similar to the segmentation approach.

\textbf{Curve-fitting approach.}
The curve-fitting approach~\cite{tabelini2021polylanenet, philion2019fastdraw} fits the lane lines into parametric curves (e.g., polynomials and splines). This approach is applied in an open-source production driver assistance system, OpenPilot~\cite{openpilot}. The main advantage of this approach is lightweight computation, allowing OpenPilot to run on a smartphone-like device without GPU. 
To achieve high efficiency, the accuracy is generally not high as other approaches. Additionally, prior work mentions that this approach is biased toward straight lines because the majority of lane lines in the training data are straight~\cite{tabelini2021polylanenet}.

\textbf{Anchor-based approach.}
Anchor-based approach~\cite{tabelini2021cvpr, li2019line, qu2021fololane} is inspired by region-based object detectors such as Faster R-CNN~\cite{ren2015faster}. In this approach, each lane line is represented as a straight proposal line (anchor) and lateral offsets from the proposal line. Similar to the row-wise classification approach, this approach takes advantage of the domain-specific knowledge that the lane lines are generally straight. This design enables to achieve state-of-the-art latency and performance. 
LaneATT~\cite{tabelini2021cvpr} reports that it achieves a higher F1 score (96.77\%) than the segmentation approaches (95.97\%) ~\cite{hou2019learning, pan2018spatial} on the TuSimple dataset.

\begin{figure*}[t!]
\vspace{-0.1in}
\centering
\includegraphics[width=0.95\linewidth]{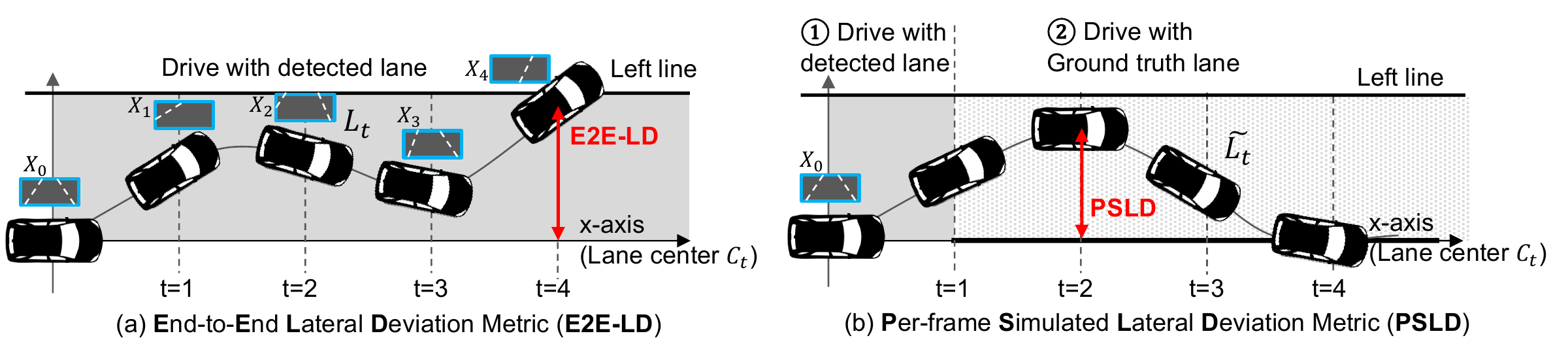}
\vspace{-0.15in}
\caption{Overview of our driving-oriented metrics for lane detection models: E2E-LD and PSLD. $X_t$ are camera frames from driver's view (lane detection model inputs). E2E-LD requires multiple (consecutive) camera frames, while PSLD only uses the current frame $X_0$.}
\label{fig:overview}
\vspace{-0.2in}
\end{figure*}

\nsubsection{Evaluation Metrics for Lane Detection} \label{sec:problem}

All lane detection methods we discuss in~\S\ref{sec:lane_detection} evaluate their performance on the \textit{accuracy} and \textit{F1 score} metrics used in the 2017 TuSimple Challenge~\cite{tusimple}. The \textit{accuracy} is calculated by $\sum_{i \in H} \frac{{\rm tp}_{i}}{|H|}$, where $H$ is a set of sampled y-axis points in the driver's view image and ${\rm tp}_{i}$ is 1 if the difference of a predicted lane line point and the ground truth point at $y=i$ is within $\alpha$ pixels; otherwise is 0. $\alpha$ is set to 20 pixels in the TuSimple Challenge. The detected lane line is associated with a ground truth line with the highest accuracy.
In other datasets~\cite{pan2018spatial, llamas2019}, IoU (Intersection over Union) is also used instead of accuracy. However, the ground-truth area is only defined as a 30-pixel wide line based on lane points, and this metric is almost equivalent to accuracy.
The \textit{F1 score} is a common metric to measure the performance of binary classification tasks. This is the harmonic mean of precision and recall: $\frac{2}{{\rm recall}^{-1} + {\rm precision}^{-1}}$. In the TuSimple Challenge, the precision and recall are calculated at the lane line level: The precision is the true positive ratio of detected lane lines and the recall is the true positive ratio of ground truth lines. The true positive is defined if the accuracy of a pair of the ground truth line and detected line is $\geq$ $\beta$. $\beta$ is set to 0.85 in the TuSimple Challenge.
Although the \textit{accuracy} and \textit{F1 score} can measure the capability of lane detection at a certain level, these metrics do not fully represent the performance in the main real-world downstream application, AD~\cite{tesla2020autopilot, gmsupercruise, TANG2021107623}, as concretely shown later in~\S\ref{sec:end-to-end}.

Specifically, to reflect its performance if used in AD, or \textit{drivability}, accuracy and F1 score metrics have 2 major limitations:
\textbf{(1)} There is no justification of $\alpha = 20$ pixels and $\beta=0.85$ accuracy thresholds. For example, the ALC system can keep at the lane center even if the detection error is more than 20 pixels, as long as the detected lane lines are \textit{parallel} with actual lane lines. Furthermore, the importance of detected lane line points should not be equal, i.e., closer points to the vehicle should be more important than the distanced points to control a vehicle.
\textbf{(2)} The current metrics treat all lane lines in the driver's view equally, e.g., detection errors for the ego lane's left line are treated the same as the detection errors for the left lane's left line. However, the former is much more important to ALC systems than the latter, as the former can directly impact the downstream calculation of the lane center. For example, if a model cannot detect the left lane's left line but can still detect the ego lane's left line, it won't affect its use for ALC. However, if it cannot detect the latter but can detect the former, the accuracy metric remains the same but the downstream modules in ALC may consider the left lane's left line as ego lane's left line and thus mistakenly deviate to the left.

\nsubsection{Automated Lane Centering} \label{sec:alc}

Automated Lane Centering (ALC) is a Level-2 driving automation technology that automatically steers a vehicle to keep it centered in the traffic lane~\cite{sae2018}. Recently, ALC is widely adopted in various vehicle models such as Tesla~\cite{tesla2020autopilot} and thus one of the most popular downstream applications of lane detection. Typical ALC systems~\cite{simlink2020lane, lee2012unified, openpilot} operate in 3 modules: lane detection, lateral control, and vehicle actuation. \newpart{More details of ALC are in Appendix~\ref{appendix:openpilot}.} While there is a line of research that designs end-to-end DNNs for ALC or higher driving automation~\cite{casas2021mp3, bojarski2016end, bansal2019chauffeurnet}, the current industry-standard solutions adopt such a modular design to ensure accountability and safety.
In the lateral control, ALC plans to follow the lane center as waypoints with Proportional-Integral-Derivative (PID)~\cite{dorf2011modern} or Model Predictive Control (MPC) \cite{MPC}.

\textbf{Adversarial Attack on ALC.}
After researchers found DNN models generally vulnerable to adversarial attacks~\cite{Szegedy2014, goodfellow2014explaining}, the following work further explored such attacks in the physical world~\cite{brown2017advpatch, eykholt2018physical}. A recent study demonstrates that ALC systems are also vulnerable to physical-world adversarial attacks~\cite{sato2020hold}. Their attack, dubbed Dirty Road Patch (DRP) attack, targets industry-grade DNN-based ALC systems, and is designed to be robust to the vehicle position and heading changes caused by the attack in the earlier frames. We use the DRP attack to simulate challenging but realizable scenarios in our evaluations.

%% file: src/methodology.tex
\nsection{Methodology}\label{sec:methodology}

In this section, we motivate the design of 2 new downstream task-aware metrics to measure the performance of lane detection models in ALC. To evaluate the validity of the metrics even in challenging scenarios, we formulate attack objective functions to fairly generate adversarial attacks against the 4 major types of lane detection methods.

\subsection{End-to-End Lateral Deviation Metric}\label{sec:E2ELD}

As the name of ALC indicates, the performance of ALC should be evaluated by how accurately it can drive in the lane center, i.e., the \textit{lateral} (left or right) deviation from the lane center. In particular, the maximum lateral deviation from the lane center in continuous closed-loop perception and control is the ultimate downstream-task performance metric for lane detection. Such deviation is directly safety-critical as large lateral deviations can cause a fatal collision with other driving vehicles or roadside objects. We call it \textit{End-to-End Lateral Deviation metric} \textbf{(E2E-LD)}, shown in Fig.~\ref{fig:overview} (a). The E2E-LD at $t=0$ is obtained as follows.
\vspace{-0.02in}
\begin{align}
    \max_{t \leq T_E}(|L_t - C_t|)
\end{align},
where $L_t$ is the lateral (y-axis) coordinate of the vehicle at $t$. $C_t$ is the lane center lateral (y-axis) coordinate corresponding to the vehicle position at $t$. We use the vehicle coordinate system at $t=0$. $T_E$ is a hyperparameter to decide the time duration. If $T_E = $ 1 second, the E2E-LD is the largest deviation within one second. 
To obtain $L_t$, it requires a closed-loop mechanism to simulate a driving by ALC, such as AD simulators~\cite{carla, lgsvl}. Starting from $t=0$, the vehicle position and heading at $t=1$ is calculated based on the camera frame at $t=0$ ($X_0$): The lane detection model detects lane lines from the frame, the lateral control interprets it by a steering angle, and vehicle actuation operates the steering wheel. This procedure repeats until $t=T_e$. Hence, \textit{multiple} (consecutive) camera frames $X_0,...,X_{T_E}$ are required and they are dynamically changed based on the lane detection results in the earlier frames.

However, such AD simulations are too computationally expensive for large-scale evaluations. Thus, we simulate vehicle trajectories by following prior work~\cite{sato2020hold}, which combines vehicle motion model~\cite{bicyclemodel} and perspective transformation~\cite{hartley2003perspective, tanaka2011perspective} to dynamically synthesize camera frames from existing frames according to a driving trajectory.

\subsection{Per-Frame Simulated Lateral Deviation Metric}

The E2E-LD metric is defined as the desired metric based on the requirements of downstream task ALC. However, it is still too computationally intensive to be monitored during training of the upstream lane detection model.
This overhead is mainly due to the camera frame inter-dependency that the camera frames are dynamically changed based on the lane detection results in the earlier frames.
To address this limitation, we design the \textit{Per-Frame Simulated Lateral Deviation metric} \textbf{(PSLD)}, which simulates E2E-LD only with a \textit{single} camera input at the current frame ($X_0$) and the geometry of the lane center. 

The overview of PSLD is shown in Fig.~\ref{fig:overview} (b). The calculation consists of two stages: \circled{1} update the vehicle position with the current camera frame at $t=0$ ($X_0$) and its lane detection result, and \circled{2} apply the closed-loop simulation using the ground-truth lane center as waypoints from $t=1$ to $t=T_p$. Note that we do not need camera frames in \circled{2} as the vehicle just tries to follow the ground-truth waypoints with lateral control, i.e., we bypass the lane detection assuming we know the ground-truth in $t\geq 1$.
We then take the maximum lateral deviation from the lane center as a metric as with E2E-LD. For convenience, we normalize the maximum lateral deviation by $T_p$ to make it a \textit{per-frame} metric. The definition of PSLD is as follows:
\vspace{-0.05in}
\begin{align}
    \frac{1}{T_p}\max_{1 \leq t \leq T_p}(|\widetilde{L}_t - C_t|)
\end{align}
, where the $\widetilde{L}_t$ is the simulated lateral (y-axis) coordinate of the vehicle at $t$. For example, for $T_p=1$, it is just a single-step simulation with the current lane detection result. The longer $T_p$ can simulate the tailing effect of the current frame result in the later frames, but it  may suffer from accumulated errors.
In~\S\ref{sec:PSLD_eval}, we explore which $T_p$ achieves the best correlation between PSLD and E2E-LD. More details are in Appendix~\ref{appendx:metric}.

\vspace{-0.05in}
\subsection{Attack Generation} \label{sec:attack_gen}
\vspace{-0.05in}

In this study, we utilize and extend physical-world adversarial attacks to evaluate the robustness of the lane detection system against challenging but realizable scenarios. To fairly generate adversarial attacks for all 4 major types of lane detection methods, we design an attack objective that can be commonly applicable to them. We name it the \textit{expected road center}, which averages all detected lane lines weighted with their probabilities. Intuitively, the average of all lane lines is expected to represent the road center. If the expected center locates at the center of the input image, its value will be 0.5 in the normalized image width. We maximize the expected road center to attack to the right and minimize it to attack to the left. Detailed calculation of the expected road center for each method is as follows.

\textbf{Segmentation \& row-wise classification approaches:}
\vspace{-0.15in}
\begin{align}
    \frac{1}{L\cdot H}\sum_{l = 1}^{L}\sum_{i = 1}^{W}\sum_{j = 1}^{H} i \cdot P^l_{ij}
\end{align}
, where $H$ and $W$ are the height and width of probability map, $L$ is the number of probability maps (channels), and $P^l_{ij}$ is the lane line existence probability of the pixel in the $(i, j)$ element of the probability map. %

\textbf{Curve-fitting approach:}
\vspace{-0.03in}
\begin{align}
   \frac{1}{L\cdot |\mathcal{H}|}\sum_{l = 1}^{L}\sum_{j \in \mathcal{H}} [j^d, j^{d-1}, \cdots, j, 1] p_l
\end{align}
, where $L$ is the number of detected lane lines, $d$ is the degrees of polynomial ($d=3$ used in PolyLaneNet~\cite{tabelini2021polylanenet}), $\mathcal{H}$ is a set of sampled y-axis values, and $p_l\in \mathbb{R}^{d + 1}$ is the coefficient of detected lane line $l$.

\textbf{Anchor-based approach:}
\vspace{-0.03in}
\begin{align}
    \sum_{l \in \mathcal{A}}
    \left [ \frac{1}{|\Delta^l|} \sum_{j \in \Delta^l}
    (a^l_j + \delta^l_j) \right ]  \cdot \pi^l
\end{align}
, where $\mathcal{A}$ is a set of the anchor proposals, $\Delta^l$ is an index set of y-axis value for anchor proposal $l$, $\pi^l$ is the probability of anchor proposal $l$, and
$a^l_j$ and $\delta^l_j$ are the x-axis value and its offset of anchor proposal $l$ at y-axis index $j$ respectively.

We incorporate this expected road center functions into DRP attack~\cite{sato2020hold} procedure to generate adversarial attacks that are effective for multiple frames.

%% file: src/experiments.tex
\nsection{Experiments} \label{sec:experiment}

\begin{figure*}[t!]
\centering
\includegraphics[width=0.95\linewidth]{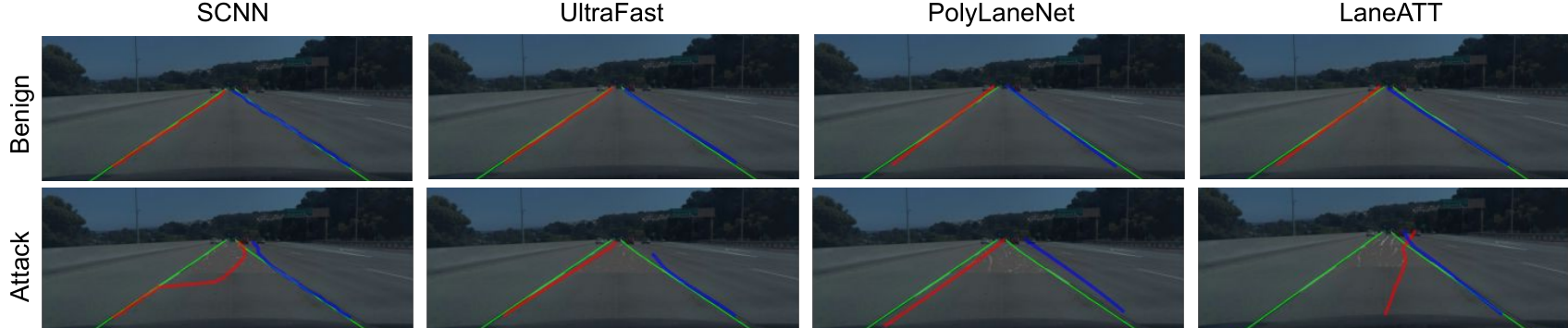}
\vspace{-0.1in}
\caption{Examples of the benign and attack-to-the-right scenarios on the Comma2k19-LD dataset. The red, blue, and green lines are the detected left and right lines and the ground-truth lines respectively.}
\label{fig:poc_comma}
\vspace{-0.2in}
\end{figure*}

We conduct a large-scale empirical study to evaluate the validity of the conventional metrics and our PLSD by comparing them with the ultimate downstream-task performance metric E2E-LD. We evaluate the 4 major types of lane detection approaches. We select a representative model for each approach as shown in Table~\ref{tbl:lane_detection}. The pretrained weights of all models are obtained from the authors' or publicly available websites\footnote{\notsotiny
LaneATT {\notsotiny \url{https://github.com/lucastabelini/LaneATT}}\\
SCNN {\notsotiny \url{https://github.com/harryhan618/SCNN_Pytorch}}\\
UltraFast {\notsotiny \url{https://github.com/cfzd/Ultra-Fast-Lane-Detection}}\\
PolyLaneNet {\notsotiny \url{https://github.com/lucastabelini/PolyLaneNet}}
}.
All pretrained weights are trained on the TuSimple Challenge training dataset~\cite{tusimple}. 

\begin{table}[t]
\footnotesize
\centering
\caption{Target lane detection methods. \textit{Acc.} is the accuracy of the TuSimple Challenge dataset~\cite{tusimple} in the reference papers.}
\vspace{-0.1in}
\begin{tabular}{llll} \toprule
Approach & Selected Method & Acc.\\ \hline
Segmentation & SCNN~\cite{pan2018spatial} & 96.53\% \\ \hline
Row-wise classification & UltraFast (ResNet18)~\cite{qin2020ultra} & 95.87\% \\ \hline
Curve-fitting & PolyLaneNet (b0)~\cite{tabelini2021polylanenet} & 88.62\%\\ \hline
Anchor-based & LaneATT (ResNet34)~\cite{tabelini2021cvpr} & 95.63\%\\ \toprule
\end{tabular}
\label{tbl:lane_detection}
\vspace{-0.22in}
\end{table}

\vspace{-0.02in}
\nsubsection{Conventional Evaluation on TuSimple Dataset}\label{sec:static_eval}

\textbf{Evaluation Setup.} We first evaluate the lane detection models with the conventional accuracy and F1 score metrics on the TuSimple dataset\cite{tusimple}, which has 2,782 one-second-long video clips as test data. Each clip consists of 20 frames, and only the last frame is annotated and used for evaluation. We randomly select 30 clips from the test data. For each clip, we consider two attack scenarios: attack to the left, and to the right. Thus, in total, we evaluate 60 different attack scenarios. In each scenario, we place 3.6 m x 36 m patches 7 m away from the vehicle as shown in Fig.~\ref{fig:poc_tusimple}. To know the world coordinate, we manually calibrate the camera matrix based on the size of lane width and lane marking.
To deal with the limitation (2) discussed in ~\S\ref{sec:problem}, we remove lane lines other than the ego-left and ego-right lane lines to evaluate the applicability to ALC systems more correctly. 
\newpart{More details of each attack implementation and parameters are Appendix~\ref{appendx:input_adapt}.}

\textbf{Results.}
Table~\ref{tbl:tusimple} shows the accuracy and F1 score metrics in the benign and attacks scenarios. In the benign scenarios, LaneATT has the best accuracy (94\%) and F1 score (88\%). SCNN and UltraFast show also high accuracy and F1 score while UltraFast has the lowest F1 score (8\%) in the attack scenarios. PolyLaneNet has lower accuracy and F1 score than the others in both benign and attack scenarios.
These results are generally consistent with the reported performance as in Table~\ref{tbl:lane_detection}. However, when we visually look into the detected lane lines under attack, we find quite some cases suggesting vastly different conclusions if used in AD as the downstream task. For example, as shown in Fig.~\ref{fig:poc_tusimple}, even though SCNN has the highest accuracy in all three scenarios, its detected lane lines are heavily curved by the attack. In contrast, the detection of PolyLaneNet looks like the most robust among the 4 models, as the detected lane lines are generally parallel to the actual lane lines. However, its accuracy (63\%) is smaller than the one of SCNN (51\%) in the attack to the right scenario. \newpart{In the benign scenario, PolyLaneNet has a lower accuracy (16\% margin) than the others, but it is hard to find meaningful differences for humans as the detected lines are well-aligned with actual lane lines. We provide more examples in Appendix~\ref{appendix:openpilot}.}
Hence, the conventional accuracy and F1 score-based evaluation may not be well suitable to judge the performance of the lane detection model in representative downstream tasks such as AD.

\begin{table}[t!]
\footnotesize
\centering
\caption{Accuracy and F1 scores for attack and benign cases on the TuSimple Challenge dataset. The metrics are calculated only with ego left and right lanes. The \textbf{bold} and \underline{underlined} letters mean the highest and lowest scores, respectively, among the 4 lane detection methods. The higher score means the higher performance. %
\vspace{-0.1in}}
\label{tbl:tusimple}. 
\begin{tabular}{lccccc} \toprule
 & \multicolumn{2}{c}{Accuracy} &  & \multicolumn{2}{c}{F1 Score} \\ \cline{2-3} \cline{5-6} 
 & Benign & Attack & & Benign & Attack \\ \cline{1-6}
SCNN~\cite{pan2018spatial} & 89\% & \textbf{58\%} &  & 75\% & 28\% \\
UltraFast~\cite{qin2020ultra} & 87\% & \underline{36\%} &  & 77\% & \underline{8\%} \\
PolyLaneNet~\cite{tabelini2021polylanenet} & \underline{72\%} & 53\% &  & \underline{50\%} & 19\%\\ %
LaneATT~\cite{tabelini2021cvpr} & \textbf{94\%} & 51\%  &  & \textbf{88\%} & \textbf{29\%} \\
\toprule
\end{tabular}
\vspace{-0.25in}
\end{table}

\begin{table*}[t!]
\footnotesize
\caption{Evaluation results of the E2E-LD and the conventional metrics, accuracy and F1 in the benign and attack scenarios. For each metric, the corresponding Pearson correlation coefficient with E2E-LD in the bottom rows. The original parameters are the ones used in the TuSimple challenge. The best parameters are those that have the highest correlation between E2E-LD with respect to F1 score.
The \textbf{bold} and \underline{underlined} letters indicate the highest and lowest performance or correlation, respectively.
}
\vspace{-0.1in}
\label{tbl:summary_old_metric}
\centering
\setlength{\tabcolsep}{4.8pt}
\begin{tabular}{clccccccccccc}
\hline
&            &        & \multicolumn{4}{c}{Benign}        &  &        & \multicolumn{4}{c}{Attack}        \\ \cline{3-7} \cline{9-13} 
 & &
   &
  \multicolumn{2}{c}{\begin{tabular}[c]{@{}c@{}}Original Parameters\\ ($\alpha=20, \beta=0.85$)\end{tabular}} &
  \multicolumn{2}{c}{\begin{tabular}[c]{@{}c@{}}Best Parameters\\ ($\alpha=5, \beta=0.9$)\end{tabular}} &
   &
   &
  \multicolumn{2}{c}{\begin{tabular}[c]{@{}c@{}}Original Parameters\\ ($\alpha=20, \beta=0.85$)\end{tabular}} &
  \multicolumn{2}{c}{\begin{tabular}[c]{@{}c@{}}Best Parameters\\ ($\alpha=50, \beta=0.65$)\end{tabular}} \\ \cline{4-7} \cline{10-13} 
&            & E2E-LD [m] & Accuracy & F1   & Accuracy & F1   &  & E2E-LD [m] & Accuracy & F1   & Accuracy & F1   \\ \cline{1-7} \cline{9-13} 
\parbox[t]{2.2mm}{\multirow{4}{*}{\rotatebox[origin=c]{90}{Metric}}}  &
SCNN~\cite{pan2018spatial}&   \underline{0.21}&
  \textbf{0.93}&
  \textbf{0.84}&
  \textbf{0.59}&
  0.03&  & 
  0.48&
  \textbf{0.68}&
  \textbf{0.31}&
  \textbf{0.83}&
  0.76 \\
&UltraFast~\cite{qin2020ultra}  &  \textbf{0.18}&
  0.92&
  0.81&
  0.55&
  \textbf{0.10}&  &   0.58&
  0.60&
  0.21&
  0.82&
  \textbf{0.77} \\
&PolyLaneNet~\cite{tabelini2021polylanenet} &   0.20&
  \underline{0.78}&
  \underline{0.50}&
  \underline{0.44}&
  \underline{0.01}&  &   \textbf{0.38}&
  0.59&
  \underline{0.13}&
  \underline{0.81}&
  0.76 \\
&LaneATT~\cite{tabelini2021cvpr}&   \underline{0.21}&
  0.89&
  0.75&
  0.54&
  0.06&  &
    \underline{0.72}&
  \underline{0.51}&
  0.14&
  \underline{0.66}&
  \underline{0.48}  \\ \cline{1-8} \cline{9-13} 
\parbox[t]{2.2mm}{\multirow{4}{*}{\rotatebox[origin=c]{90}{Corr.}}} &SCNN~\cite{pan2018spatial}
&    - &
   \underline{-0.65}$^{***}$ &
   \underline{-0.60}$^{***}$ &
   \textbf{-0.33}$^{***}$ &
   -0.13$^{ns}$ &  & 
   - &
   -0.13$^{ns}$ &
   \textbf{-0.06}$^{ns}$ &
   -0.14$^{ns}$ &
   -0.06$^{ns}$ \\
&UltraFast~\cite{qin2020ultra}
&    - &
   -0.58$^{***}$ &
   -0.59$^{***}$ &
   -0.38$^{***}$ &
   \underline{-0.24}$^{*}$
  &  &    - &
   -0.24$^{*}$ &
   -0.14$^{ns}$ &
   -0.20$^{*}$ &
   \underline{-0.13}$^{ns}$ \\
&PolyLaneNet~\cite{tabelini2021polylanenet}
&    - &
   -0.60$^{***}$ &
   \textbf{-0.55}$^{***}$ &
   \underline{-0.46}$^{***}$ &
   \textbf{0.10$^{ns}$}
  &  &    - &
   \underline{-0.27}$^{**}$ &
   \underline{-0.28}$^{**}$ &
   -0.06$^{ns}$ &
   \textbf{0.01}$^{ns}$ \\
&LaneATT~\cite{tabelini2021cvpr}
&    - &
   \textbf{-0.57}$^{***}$ &
   -0.58$^{***}$ &
   -0.34$^{***}$ &
   -0.14$^{ns}$ &  &
     - &
   \textbf{0.08}$^{ns}$ &
   -0.09$^{ns}$ &
   \textbf{0.11}$^{ns}$ &
   0.12$^{ns}$
  \\ \hline
\end{tabular}
\raggedleft
$^{ns}$ Not Significant $(p > 0.05), ^{*} p \leq 0.05, ^{**} p \leq 0.01, ^{***} p \leq 0.001$
\vspace{-0.22in}
\end{table*}

\vspace{-0.02in}
\nsubsection{Consistency of TuSimple Metrics with E2E-LD}
\label{sec:end-to-end}
To more systematically evaluate the consistency of the conventional accuracy and F1 score with the performance in AD as the downstream tasks, we conduct a large-scale empirical study on our newly-constructed dataset. 

\textbf{New Dataset: Comma2k19-LD.}
To evaluate both the conventional metrics and the downstream task-centric metrics E2E-LD and PSLD on the same dataset, we need both lane line annotations and driving information (e.g., position, steering angle, and velocity). Unfortunately, there is no existing dataset that satisfies the requirements to our best knowledge. Thus, we create a new dataset, coined \textit{Comma2k19-LD}, in which we manually annotate the left and right lane lines for 2,000 frames (100 scenarios of 1-second clips at 20 Hz). The selected scenarios are randomly selected from the scenarios with more than 30 
mph ($\approx$ 48 km/h) in the original Comma2k19 dataset~\cite{comma2k19}. Fig~\ref{fig:poc_comma} shows the example frames of the Comma2k19-LD dataset. These frames are the first frames of the scenario. The following 20 frames are also annotated and the same patch is used for each attack. More details are in Appendix~\ref{appendx:comma2kld}. \newpart{The Comma2k19-LD dataset is published on our website\cite{Comma2k19-LD}.}

\textbf{Evaluation Setup.}
We conduct the evaluation on the Comma2k19-LD dataset. For the attack generation, we attack to the left in randomly selected 50 scenarios and attack to the right in the other 50 scenarios.
For the lateral control, we use the implementation of MPC~\cite{MPC} in OpenPilot v0.6.6, which is an open-source production ALC system. For the longitudinal control, we used the velocity in the original driving trace. For the motion model, we adopt the kinematic bicycle model~\cite{kong2015kinematic}, which is the most widely-used motion model for vehicles~\cite{coursera, kong2015kinematic,watzenig2016automated}. The vehicle parameters are from Toyota RAV4 2017 (e.g., wheelbase), which is used to collect the traces of the comma2k19 dataset. To make the model trained on the TuSimple dataset work on the Comma2k19-LD dataset, we manually adjust the input image size and field-of-view to be consistent with the TuSimple dataset. We place a 3.6 m x 36 m patch at 7 m away from the vehicle at the first frame. For the E2E-LD metric, we use $T_E$ = 20 frames (1 second). It follows the result that the average attack success time of the DRP attack is nearly 1 sec~\cite{sato2020hold}. More setup details are in Appendix~\ref{appendx:attack_detail},~\ref{appendx:input_adapt}, and~\ref{appendix:openpilot}).

\textbf{Results.}
Table~\ref{tbl:summary_old_metric} shows the evaluation results of conventional accuracy and F1 score and E2E-LD. We calculate the Pearson correlation coefficient $r$ and its $p$ value. As shown, there are \textit{substantial inconsistencies between the downstream-task performance (from the heavy-weight E2E-LD metric) and the conventional metrics}. 
In the benign scenarios, SCNN has the highest accuracy (0.59) and F1 score (0.84) under the original parameters ($\alpha=20, \beta=0.85$). However, SCNN is one of the methods with the \textit{lowest} E2E-LD (0.21), and instead UltraFast has the highest E2E-LD (0.18).
In the attack scenarios, the inconsistency is more obvious: PolyLaneNet has the highest E2E-LD (0.38), but PolyLaneNet achieves the 2nd lowest accuracy (0.59) and the highest F1 score (0.13) with the original parameters. 
Hence, the E2E-LD draws quite different conclusions from the conventional metrics. If we adopt the conventional metrics, SCNN should be preferred as the best performant model.
This is consistent with the results in Table~\ref{tbl:lane_detection} and ~\S\ref{sec:static_eval} since SCNN, UltraFast, and LaneATT show close performance in the conventional metrics (SCNN may have slight advantages in Comma2k19-LD).  
On the other hand, if we adopt E2E-LD, PolyLaneNet should be preferred since there is only a slight difference between the 4 lane detection methods in the benign scenarios and PolyLaneNet clearly outperforms the other methods in the attack scenarios.

The inconsistency between the E2E-LD and the conventional metrics can be more systematically quantified using Pearson correlation coefficient $r$. Generally, the E2E-LD and the conventional metrics have strongly \textit{negative} correlations ($r \leq$-0.55) with high statistical significance ($p \leq$ 0.001), meaning that some recent improvements in the conventional metrics may not have led to improvements in AD, but rather may have made it worse by overfitting to the metrics. SCNN, the segmentation approach, is the only one that does not use domain knowledge, e.g., lane lines are smooth lines (\S\ref{sec:lane_detection}). This high degree of freedom in the model may lead to overfitting of the human annotations with noise.

Finally, we evaluate the parameters in the conventional metrics: $\alpha$ for the accuracy and $\beta$ for F1 score. For $\alpha$, we explore every 5 pixels from 5 pixels to 50 pixels. For $\beta$, we explore every 0.05 from 0.5 to 0.9. In the benign scenarios, ($\alpha=20, \beta=0.85$) has the best correlation between the E2E-LD and F1 score. In the attack scenarios, ($\alpha=50, \beta=0.65$) has the best correlation between the E2E-LD and F1 score. However, the results are still similar to those using the original parameters: SCNN shows the highest accuracy; UltraFast has a higher F1 score than the others, but the correlation is still negative. Thus, such a naive parameter tuning does not resolve the limitations. of the conventional metrics.

\begin{figure}[t!]
\begin{center}
    \includegraphics[width=.424\linewidth]{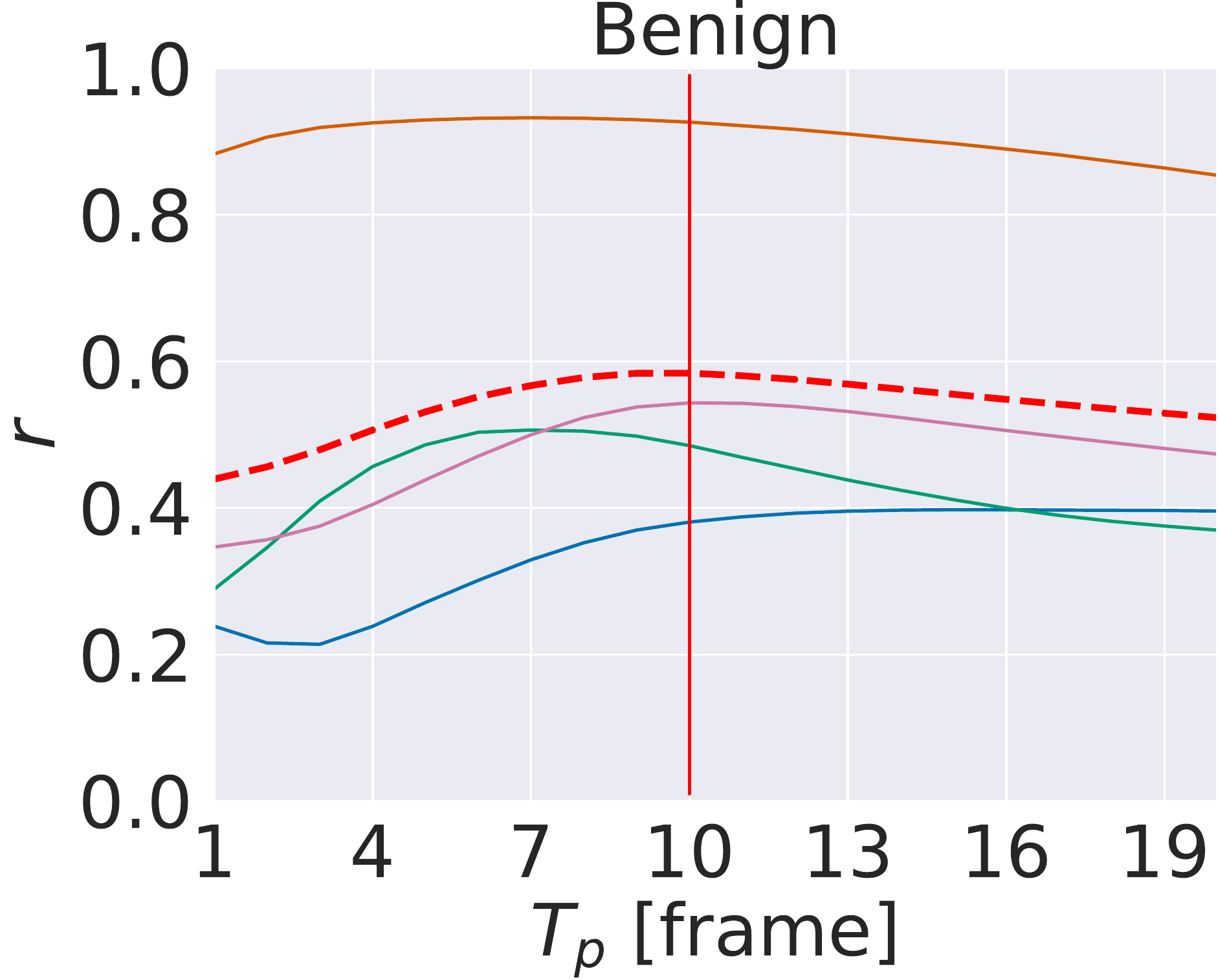}\hspace{.02\linewidth}
    \includegraphics[width=.43\linewidth]{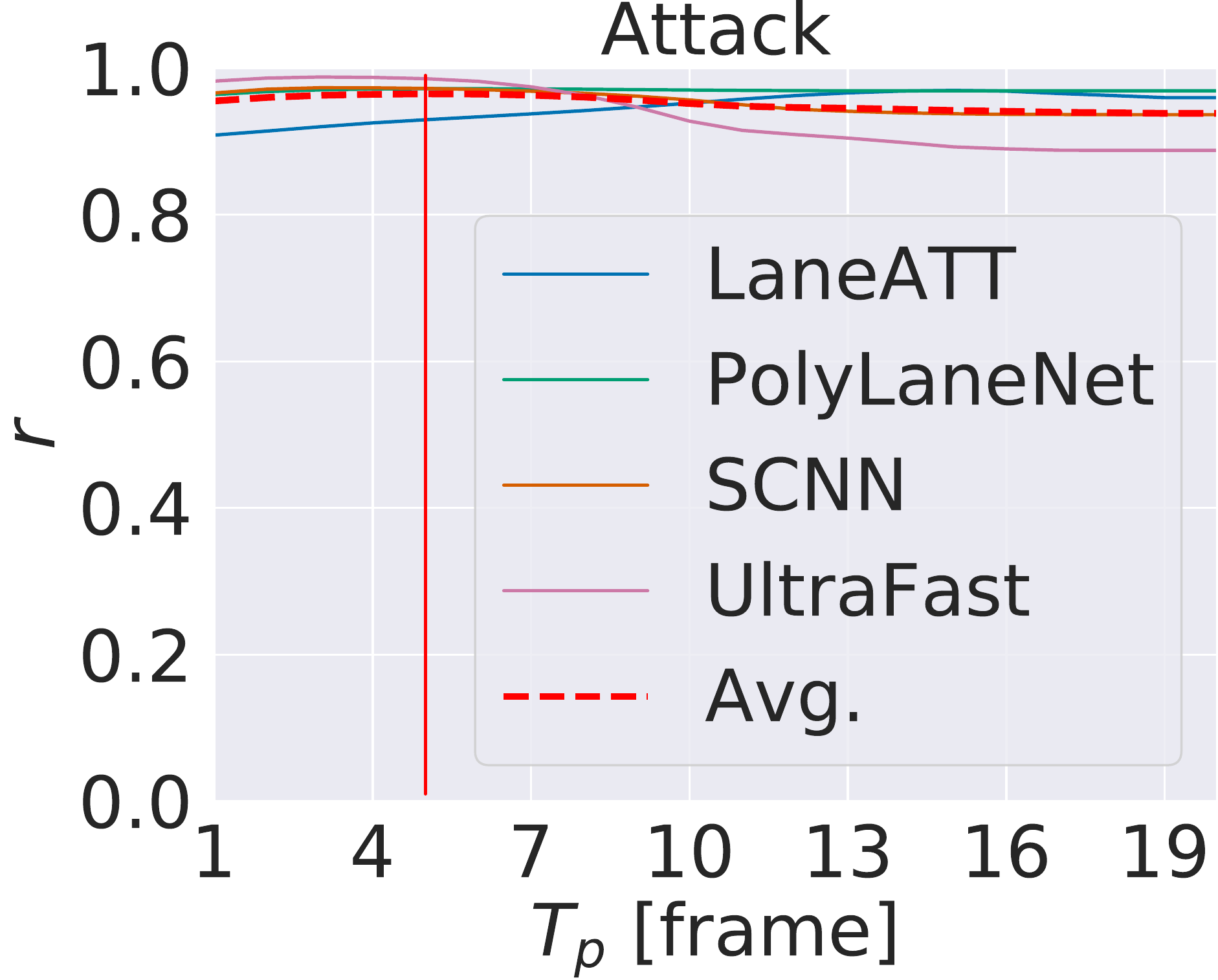}
\end{center}
\vspace{-0.22in}
\caption{Pearson correlation coefficient $r$ between E2E-LD and PSLD when $T_p$ is varied from 1 to 20 in the benign and attack scenarios. The red vertical lines are $T_p$ with the largest average $r$.}
\label{fig:PSLD_corr}
\vspace{-0.17in}
\end{figure}
\begin{figure}[t!]
\begin{center}
    \includegraphics[width=.43\linewidth]{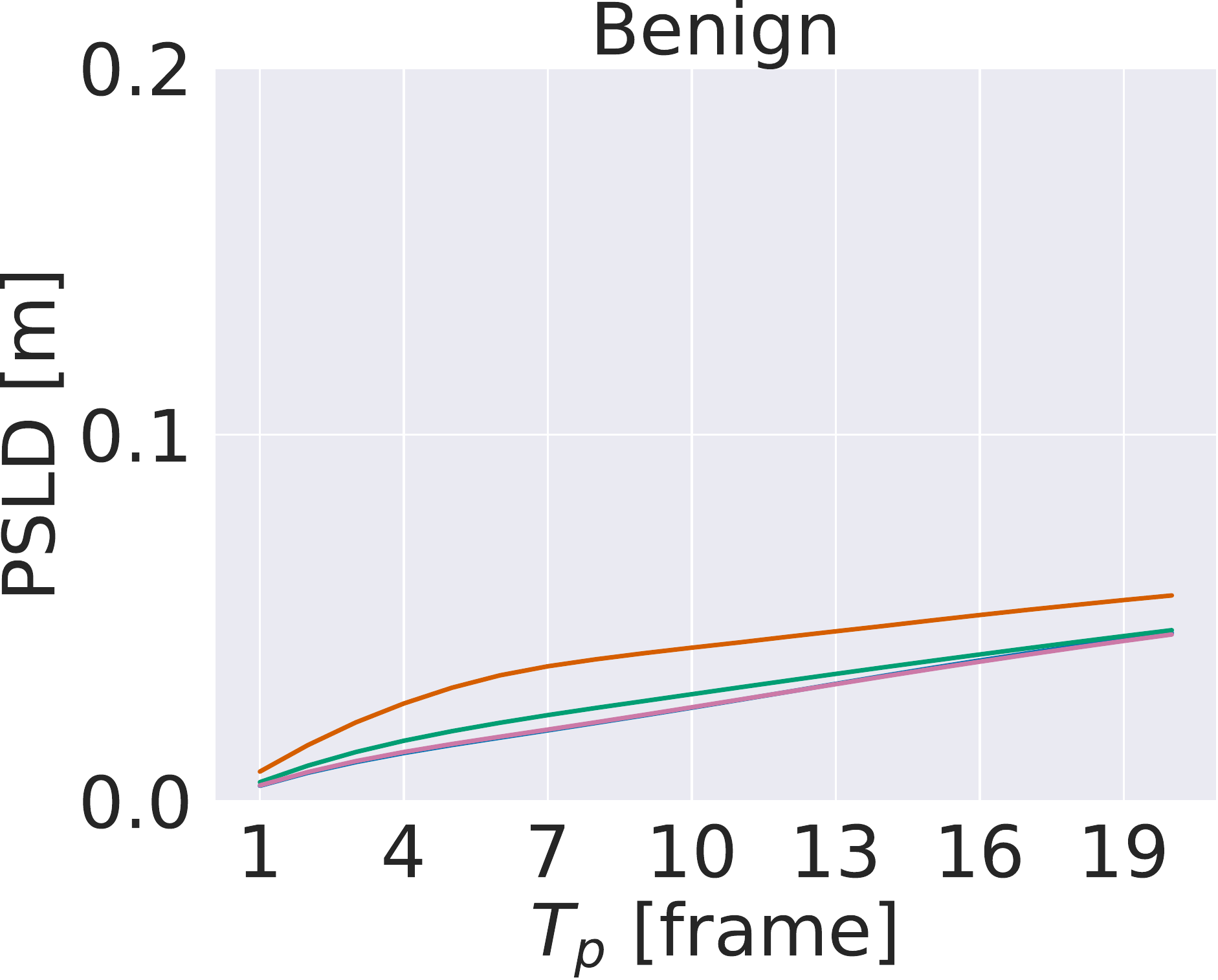}\hspace{.04\linewidth}
    \includegraphics[width=.41\linewidth]{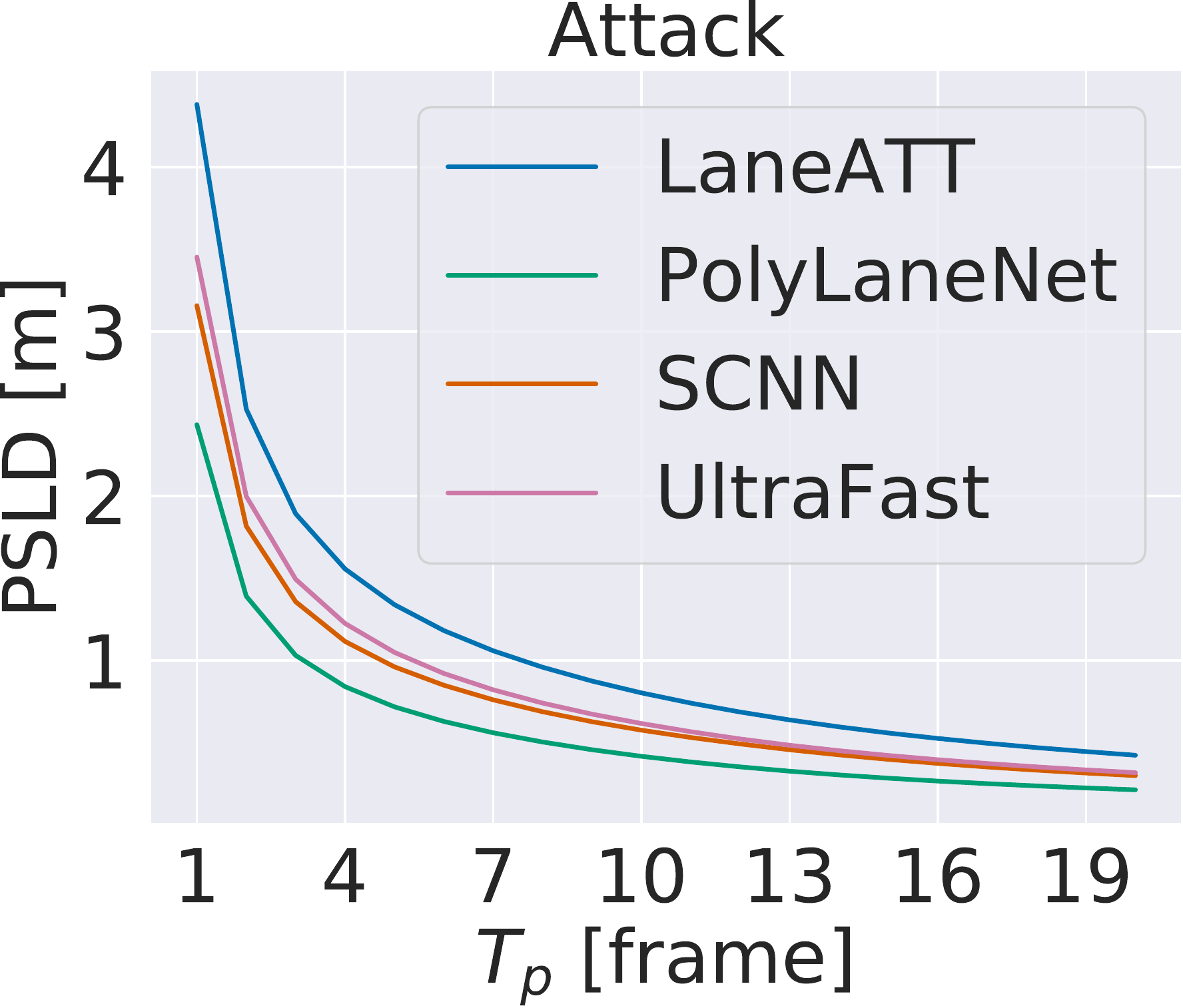}
\end{center}
\vspace{-0.23in}
\caption{PSLD for the 4 major lane detection models when $T_p$ is varied from 1 to 20 frames in the benign and attack scenarios.}
\label{fig:PSLD_values}
\vspace{-0.24in}
\end{figure}

\nsubsection{Consistency of E2E-LD with PSLD} \label{sec:PSLD_eval}

In this section, we evaluate the validity of PSLD as a \textit{per-frame} surrogate metric of E2E-LD. 

\textbf{Evaluation Setup.} 
We follow the same setup as in~\S\ref{sec:end-to-end}. We generate the DRP attacks for 100 scenarios in the Comma2k19-LD dataset with the same parameters. For the PSLD, we obtain the ground truth waypoints by the following procedure. We generate a trajectory with the bicycle model and OpenPilot's MPC by using the human driving trajectory as waypoints. We then use the generated trajectory as a ground-truth road center. While we can directly use the human driving trajectory as ground truth, human driving sometimes is not smooth and this approach can cancel the effect of motion models, which have differences from real vehicle dynamics. For the benign scenarios, we calculate the PSLD for each frame in the original human driving. For the attack scenarios, we use the frames synthesized by the method described in~\ref{sec:E2ELD} instead of the original frames because the attacked trajectory and its camera frames are largely changed from the original human driving. For example, to obtain the PSLD at frame $t=N$, we simulate the trajectory until $t=N-1$ and we then calculate the PSLD with the synthesized frame at $t=N$.

\textbf{Results.} 
Fig.~\ref{fig:PSLD_corr} shows the Pearson correlation coefficient $r$ between E2E-LD and PSLD when $T_p$ is varied from 1 to 20 frames. As shown, the E2E-LD, PSLD has strong positive correlations in both benign and attack scenarios. In particular, there are significant correlations ($>$0.8) in the attack scenarios. This is because the direction of lateral deviation generally coincides with the attack direction. By contrast, in the benign scenarios, the vehicle drives around the road center with overshooting, and thus the direction of lateral deviation heavily depends on the initial states. Nevertheless, the PSLD has always high positive correlations with E2E-LD ($>$0.2). In particular, SCNN has strong correlations ($>$0.8) with E2E-LD in all $T_p$. 
We consider that the high correlation can be due to the segmentation approach, which is the only method among the 4 methods that does not use the domain-specific knowledge the lane lines are generally smooth (\S\ref{sec:lane_detection}). The detection of SCNN at the same location tends to be consistent across different frames, i.e., SCNN is less dependent on global information.

Finally, we explore the best $T_p$ for PSLD to proxy E2E-LD. As shown in Fig.~\ref{fig:PSLD_corr}, the average of the correlation coefficients of the 4 methods achieves the maximum at $T_p=10$ in the benign scenarios and $T_p=5$ in the attack scenarios respectively. We list the E2E-LD and PLSD with $T_p=10$ and the corresponding $r$ in Table~\ref{tbl:psld_summary}.
As shown, there are strong, statistically significant ($p \leq 0.001$) positive correlations ($\geq0.38$) between E2E-LD and PSLD in both cases. The results strongly support the fact that PSLD can measure the performance of lane detection in ALCs based solely on the single camera frame and ground-truth road center geometry. We note that the PSLD is not so sensitive to the choice of $T_p$. As shown in Fig.~\ref{fig:PSLD_values}, the magnitude relation of the 4 methods is generally consistent for all $T_p$.

\begin{table}[t!]
\centering
\footnotesize
\caption{Evaluation results of the E2E-LD and PSLD in the benign and attack scenarios. The format is the same as Table~\ref{tbl:summary_old_metric}.
}
\vspace{-0.1in}
\label{tbl:psld_summary}
\setlength{\tabcolsep}{1.6pt}
\begin{tabular}{clccccccc}
\hline
&             & \multicolumn{2}{c}{Benign}          &  & \multicolumn{2}{c}{Attack}          \\ \cline{3-4} \cline{6-7}
&             & E2E-LD [m] & PSLD [m]   &  & E2E-LD [m] & PSLD [m]   \\ \cline{1-4} \cline{6-7} 
\parbox[t]{2.2mm}{\multirow{4}{*}{\rotatebox[origin=c]{90}{Metric}}}&
SCNN~\cite{pan2018spatial}& 
\underline{0.21}&  \underline{0.04}&  &  0.48&  0.58\\ 
& UltraFast~\cite{qin2020ultra} & \textbf{0.18} & \textbf{0.03}&  &  0.58 & 0.62\\ 
& PolyLaneNet~\cite{tabelini2021polylanenet} & 
0.20 &  \textbf{0.03}&  & \textbf{0.38}& \textbf{0.42}\\
& LaneATT~\cite{tabelini2021cvpr} & \underline{0.21}&  \textbf{0.03}  &  &  \underline{0.72}&  \underline{0.80}
\\ \hline
\parbox[t]{2.2mm}{\multirow{4}{*}{\rotatebox[origin=c]{90}{Corr.}}}&
SCNN~\cite{pan2018spatial}
&  -   & 
 \textbf{0.93}$^{***}$        &  & 
 -   & 
 0.96$^{***}$        \\
& UltraFast~\cite{qin2020ultra}
&  -    & 
 0.54$^{***}$        &  & 
 -   & 
 \underline{0.93}$^{***}$        \\
& PolyLaneNet~\cite{tabelini2021polylanenet}
&  -   & 
 0.49$^{***}$        &  & 
 -   & 
 \textbf{0.97}$^{***}$ \\
& LaneATT~\cite{tabelini2021cvpr}
&  -   & 
 \underline{0.38}$^{***}$        &  & 
 -   & 
 0.95$^{***}$\\ \hline
\end{tabular}
\raggedleft
$^{ns}$ Not Significant $(p > 0.05), ^{*} p \leq 0.05, ^{**} p \leq 0.01, ^{***} p \leq 0.001$
\vspace{-0.2in}
\end{table}

%% file: src/discussion.tex
\nsection{Discussion} \label{sec:discussion}

\newpart{
\textbf{Alternative Metric Design.}
To improve the existing metrics, we explored other possible design choices. One of the most intuitive approaches is the $\mathcal{L}_{1}$ or $\mathcal{L}_{2}$ distance in the bird's eye view. 
We evaluated the designs and confirmed that these metrics are still leading to erroneous judgment on downstream AD performance similar to the conventional metrics. Details are in Appendix~\ref{appendix:3D-L1L2}. We note that our metrics are specific to AD, the main downstream task of lane detection. For other downstream tasks, other metric designs can be more suitable.
}

\textbf{Domain Shift.} In this work, we use lane detection models pretrained on the TuSimple dataset and evaluate them on the Comma2k19-LD. \newpart{To evaluate the impact of domain shift, we conduct further evaluation and confirm that our observations are generally consistent.
Detailed results and discussions are in Appendix~\ref{appendix:domain_shift}.
}

\textbf{Closed-loop Simulation.} 
To obtain driving-oriented metrics, there are multiple parameters and design choices in the closed-loop simulation. In this study, we follow the parameters in the Comma2k19 datasets and select simple and popular designs, e.g., bicycle model and MPC. Meanwhile, we think that such design differences should only have minor effects on our observations because ALC, Level-2 driving automation, just follows the lane center line, which is designed to be smooth in normal roads.

\newpart{
\textbf{Evaluation on Other Datasets.} 
Our metrics are applicable to any dataset set that contains position data (e.g. GPS) and its camera frames, but ideally, velocity and ground-truth lane centers should be available. Such information is available in relatively new datasets such as~\cite{waymo_open_dataset, chang2019argoverse}.
However, lane annotations are not directly available in the datasets and require considerable effort to obtain from map data and camera frames. To our knowledge, our Comma2k19-LD is so far the only dataset with both lane line annotation and driving information. We hope our work will facilitate further research to build datasets including them.
}

%% file: src/conclusion.tex
\nsection{Conclusion}

In this work, we design 2 new lane detection metrics, E2E-LD and PSLD, which can more faithfully reflect the performance of lane detection models in AD.
Throughout a large-scale empirical study of the 4 major types of lane detection approaches on the TuSimple dataset and our new dataset Comma2k19-LD, we highlight critical limitations of the conventional metrics and demonstrate the high validity of our metrics to measure the performance in AD, the core downstream task of lane detection.
In recent years, a wide variety of pretrained models have been used in many downstream application areas such as AD~\cite{apollo}, natural language processing~\cite{bert}, and medical~\cite{chen2019med3d}. Reliable performance measurement is essential to facilitate the use of machine learning responsibly. 
We hope that our study will help the community make further progress in building a more downstream task-aware evaluation for lane detection.

%% file: src/appendix.tex
\section{Detailed Settings of Lane Detection Metrics} \label{appendx:metric}

Table~\ref{tbl:metric_param} shows the parameters and input required to calculate each metric for lane detection. As shown, only E2E-LD requires multiple frames for computation. For E2E-LD and PSLD, we adopt the kinematic bicycle model~\cite{kong2015kinematic} to simulate the vehicle motion. The only parameter in the kinematic bicycle model is the wheelbase. We use WheelBase=2.65 meters, which is the wheelbase of Toyota RAV4 2017.

\begin{table}[h!]
\setlength{\tabcolsep}{3pt}
\caption{Parameter and input of metrics for Lane Detection}
\begin{tabular}{lccc}
\toprule
         & Parameter   & Input         & Per-frame \\ \hline
Accuracy & $\alpha$       & $X_0$            & \checkmark          \\ 
F1 score & $\alpha, \beta$ & $X_0$            & \checkmark          \\ 
E2E-LD   & $T_E$, WheelBase        & $X_0,...,X_{T_E}, C$ &           \\ 
PSLD     & $T_p$, WheelBase          & $X_0, C$         & \checkmark           \\
\toprule
\end{tabular}
\label{tbl:metric_param}
\end{table}

\section{Detailed Attack Implementation} \label{appendx:attack_detail}

We use the official implementation of the DRP attack~\cite{sato2020hold}. 
We also use parameters that are reported to have the best balance between effectiveness and secrecy: the learning rate is $10^{-2}$, the regularization parameter $\lambda$ is $10^{-3}$, and the perturbable area ratio (PAR) is 50\%. We run 200 iterations to generate the patch in all experiments.

\section{Details of Comma2k19-LD dataset} \label{appendx:comma2kld}

\begin{figure*}[htb]
\centering
\includegraphics[width=\linewidth]{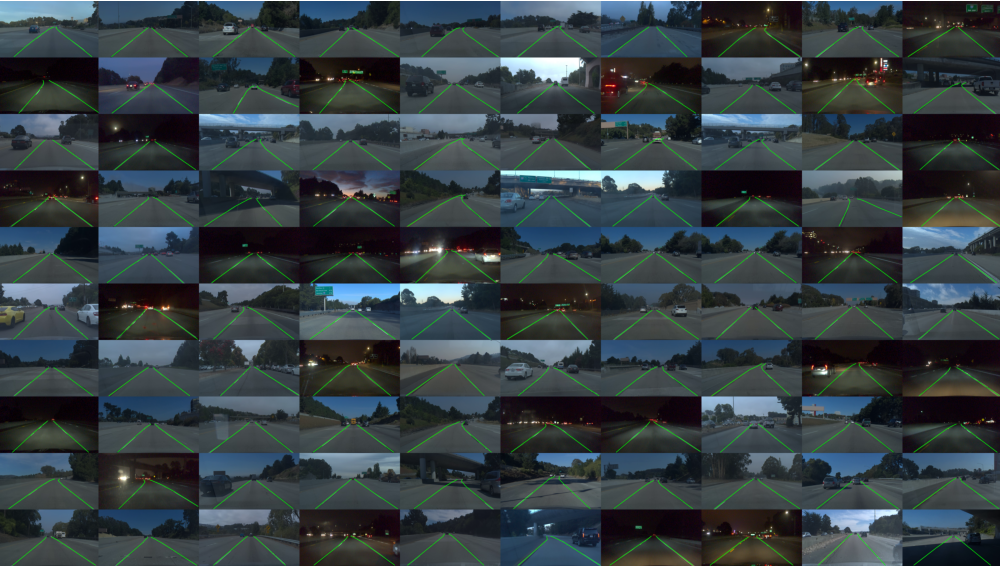}
\caption{The first frame of all 100 scenarios and its lane line annotations (green line)}
\label{fig:comma2k19ld}
\end{figure*}

Fig.~\ref{fig:comma2k19ld} shows the first frame of all 100 scenarios and its lane line annotations. For annotation, human annotators mark the lane line points and check if the linear interpolation results of the markings align with the lane line information. To convert the annotations to the TuSimple dataset format, we sample points every 10 pixels in the y-axis from the interpolated results.

\section{Adaptation to TuSimple Challenge Camera Frames Geometory} \label{appendx:input_adapt}

\begin{figure}[h]
\centering
\includegraphics[width=\linewidth]{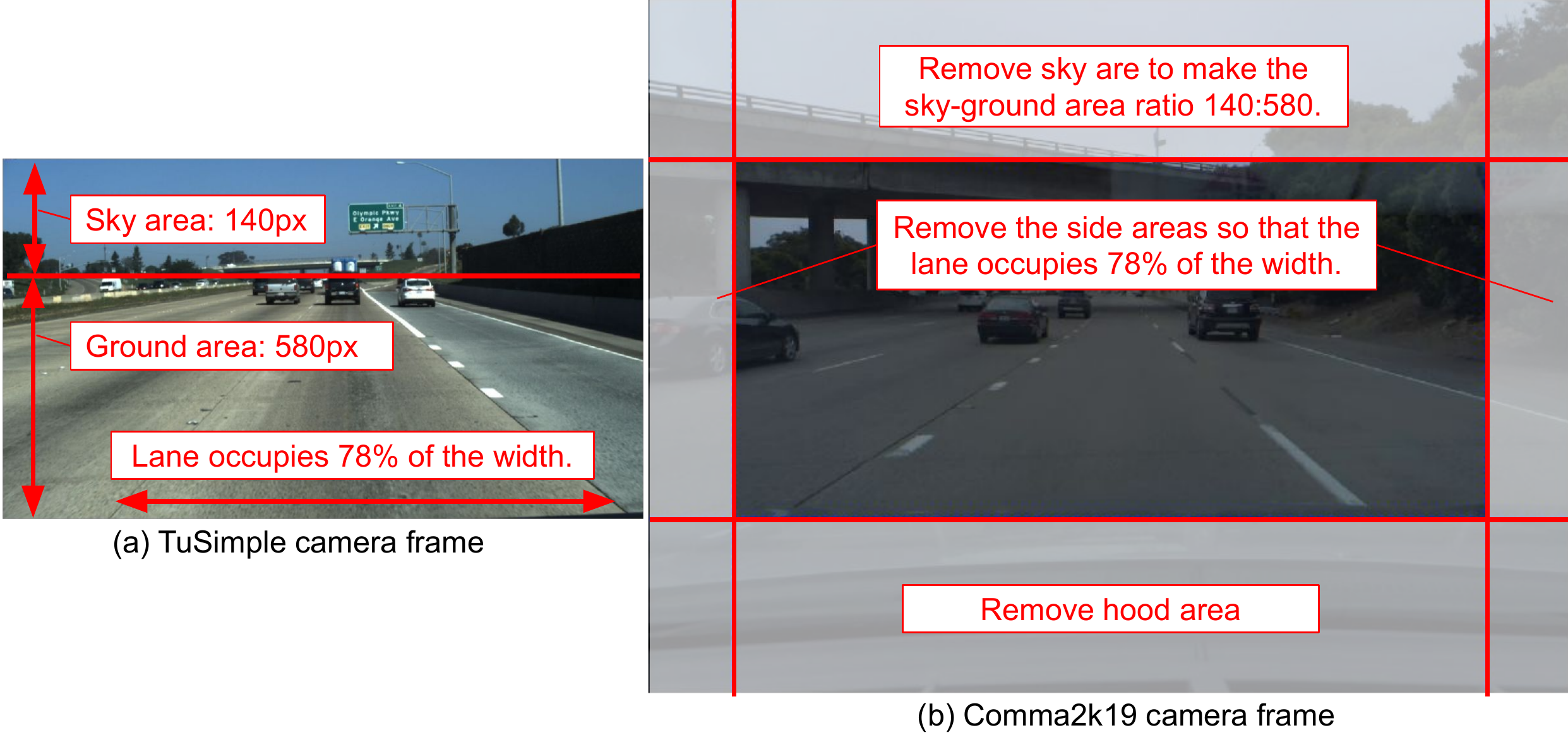}
\caption{Overview of adapting the camera frames in Comma2k19-LD dataset to the camera frame in the TuSimple Challenge dataset. We remove the surrounding area and use only the central part of the Comma2k19 camera frame to ensure that the comma2k19-LD camera frames have a geometry similar to that of the TuSimple challenge camera frames.}
\label{fig:comma2tusimple}
\end{figure}

In the evaluation of the comma2k19-LD dataset, we use the same pretrained models trained on the TuSimple Challenge training dataset. To deal with the differences in the datasets, we convert the camera frames in the comma2k19-LD dataset to have geometry similar to the camera frames in the TuSimple challenge dataset.
Fig.~\ref{fig:comma2tusimple} illustrates the overview of the conversion. We remove the surrounding area and use only the central part of the Comma2k19-LD camera frame to have the same sky-ground area ratio and the same lane occupation ratio in the image width with the ones in the TuSimple dataset. 

\section{Evaluation of the Domain Shift Effect} \label{appendix:domain_shift}

In this study, we use lane detection models pretrained on the TuSimple dataset and evaluate them on the Comma2k19-LD dataset. Although both datasets are similar driver's view images, there can be some domain shifts between them. To understand the impact, we trained the 4 models with another 100 scenarios extracted from the Comma2k19 dataset. We run 10 epochs with the data on top of the models pretrained on the TuSimple dataset. For the lane line labels, we use OpenPilot's lane detection results in the dataset. We conduct the same evaluation in~\S4.2 and~\S4.3 of the main paper. As shown in Table~\ref{tbl:summary_old_metric} and Table~\ref{tbl:psld_summary}, the observations are consistent: SCNN outperforms in the conventional metrics; PolyLaneNet is the most robust in attack scenarios. The Pearson correlation coefficients show almost the same results as the ones in~\S4.2 and~\S4.3 that the conventional metrics have strong negative correlations in the benign scenarios and the correlations in the attack scenarios are not statistically significant. 

However, the E2E-LD in the attack scenarios are generally higher than the results in~\S4.2 and~\S4.3 of the main paper while the E2E-LD in the benign scenarios is generally lower. This indicates that this additional fine-tuning improves the performance in the benign scenarios, but it harms the robustness against adversarial attacks. 

\begin{table*}[t!]
\footnotesize
\caption{Evaluation results of the E2E-LD and the conventional metrics, accuracy and F1 in the benign and attack scenarios. For each metric, the corresponding Pearson correlation coefficient with E2E-LD in the bottom rows.
The \textbf{bold} and \underline{underlined} letters indicate the highest and lowest performance or correlation, respectively.
}
\label{tbl:summary_old_metric}
\centering
\setlength{\tabcolsep}{4.9pt}
\begin{tabular}{clccccccc}
\hline
&            &        \multicolumn{3}{c}{Benign}        &  &        \multicolumn{3}{c}{Attack}        \\ \cline{3-5} \cline{7-9} 
 & &
   &
  \multicolumn{2}{c}{\begin{tabular}[c]{@{}c@{}}Original Parameters\\ ($\alpha=20, \beta=0.85$)\end{tabular}} &
   &
   &
  \multicolumn{2}{c}{\begin{tabular}[c]{@{}c@{}}Original Parameters\\ ($\alpha=20, \beta=0.85$)\end{tabular}} \\ \cline{4-5} \cline{8-9} 
&            & E2E-LD [m] & Accuracy & F1    &  & E2E-LD [m] & Accuracy & F1    \\ \cline{1-5} \cline{7-9} 
\parbox[t]{2.2mm}{\multirow{4}{*}{\rotatebox[origin=c]{90}{Metric}}}  &
SCNN~\cite{pan2018spatial}&   \underline{0.20}&
  \textbf{0.93}&
  \underline{0.84}& &
  \textbf{0.52}&
  \textbf{0.67}&
  0.30 \\
&UltraFast~\cite{qin2020ultra}  &  0.18&
  \underline{0.92}&
  \underline{0.84}
  &  &   0.62&
  \underline{0.49}&
  0.16\\
&PolyLaneNet~\cite{tabelini2021polylanenet} &   \textbf{0.13}&
  \textbf{0.93}&
  \textbf{0.86}&  &   0.54&
  0.62&
  \textbf{0.33} \\
&LaneATT~\cite{tabelini2021cvpr}&   0.14&
  \textbf{0.93}&
  0.85&  &
    \underline{0.71}&
  0.51&
  \underline{0.12}\\ \cline{1-5} \cline{7-9} 
\parbox[t]{2.2mm}{\multirow{4}{*}{\rotatebox[origin=c]{90}{Corr.}}} &SCNN~\cite{pan2018spatial}
&    - &
   \underline{-0.65}$^{***}$ &
   -0.51$^{***}$ &  & 
   - &
   \underline{-0.09}$^{ns}$ &
   -0.04$^{ns}$ \\
&UltraFast~\cite{qin2020ultra}
&    - &
   -0.63$^{***}$ &
   -0.60$^{***}$ &  &    - &
   \textbf{0.14}$^{ns}$ &
   \textbf{0.07}$^{ns}$ \\
&PolyLaneNet~\cite{tabelini2021polylanenet}
&    - &
   \textbf{-0.32}$^{***}$ &
   \underline{-0.62}$^{***}$ &  &    - &
   \textbf{0.14}$^{ns}$ &
   0.04$^{ns}$  \\
&LaneATT~\cite{tabelini2021cvpr}
&    - &
   -0.57$^{***}$ &
   \textbf{-0.26}$^{***}$ &  &
     - &
   -0.02$^{ns}$ &
   \underline{-0.06}$^{ns}$ \\ \hline
\end{tabular}

\raggedleft
$^{ns}$ Not Significant $(p > 0.05), ^{*} p \leq 0.05, ^{**} p \leq 0.01, ^{***} p \leq 0.001$ \hspace{9em}
\end{table*}

\begin{table}[t!]
\centering
\footnotesize
\caption{Evaluation results of the E2E-LD and PSLD in the benign and attack scenarios. The format is the same as Table~\ref{tbl:summary_old_metric}.
}
\label{tbl:psld_summary}
\setlength{\tabcolsep}{1.6pt}
\begin{tabular}{clccccccc}
\hline
&             & \multicolumn{2}{c}{Benign}          &  & \multicolumn{2}{c}{Attack}          \\ \cline{3-4} \cline{6-7}
&             & E2E-LD [m] & PSLD [m]   &  & E2E-LD [m] & PSLD [m]   \\ \cline{1-4} \cline{6-7} 
\parbox[t]{2.2mm}{\multirow{4}{*}{\rotatebox[origin=c]{90}{Metric}}}&
SCNN~\cite{pan2018spatial}& 
\underline{0.20}&  \underline{0.04}&  &  \textbf{0.52}&  0.61\\ 
& UltraFast~\cite{qin2020ultra} & 0.18 & \textbf{0.02}&  &  0.62 & 0.66\\ 
& PolyLaneNet~\cite{tabelini2021polylanenet} & 
\textbf{0.13} &  \textbf{0.02}&  & 0.54& \textbf{0.55}\\
& LaneATT~\cite{tabelini2021cvpr} & 0.14&  0.03  &  &  \underline{0.71}&  \underline{0.82}
\\ \hline
\parbox[t]{2.2mm}{\multirow{4}{*}{\rotatebox[origin=c]{90}{Corr.}}}&
SCNN~\cite{pan2018spatial}
&  -   & 
 \textbf{0.93}$^{***}$        &  & 
 -   & 
 0.93$^{***}$        \\
& UltraFast~\cite{qin2020ultra}
&  -    & 
 0.60$^{***}$        &  & 
 -   & 
 \textbf{0.99}$^{***}$        \\
& PolyLaneNet~\cite{tabelini2021polylanenet}
&  -   & 
 0.65$^{***}$        &  & 
 -   & 
 \textbf{0.99}$^{***}$ \\
& LaneATT~\cite{tabelini2021cvpr}
&  -   & 
 \underline{0.55}$^{***}$        &  & 
 -   & 
 \underline{0.78}$^{***}$\\ \hline
\end{tabular}
\raggedleft
$^{ns}$ Not Significant $(p > 0.05), ^{*} p \leq 0.05, ^{**} p \leq 0.01, ^{***} p \leq 0.001$
\end{table}

\section{Alternative metric design} \label{appendix:3D-L1L2}
\newpart{
To improve the conventional metrics, one of the most intuitive approaches is the $\mathcal{L}_{1}$ or $\mathcal{L}_{2}$ distance in the bird's eye view because they do not suffer from the problem of the ill parameters discussed in~\S2.2, and lane detection results from a bird's eye view may be a more adequate to measure of drivability than detection results from a front camera. 
We actually have considered such metrics before, but we did not finally choose them because, without some form of control simulation, we find it fundamentally nontrivial to accurately predict the combined effects of detection errors at different lane line positions and with different error amounts on the downstream AD driving. This can be concretely shown in Table~\ref{tbl:3D-L1L2}.
As shown, both such 3D-L1 and 3D-L2 distance metrics have considerably lower correlation coefficient $r$ with E2E-LD compared to our PSLD. They are indeed better than conventional accuracy and F1 score metrics. However, they are still leading to erroneous judgment on downstream AD performance similar to the accuracy and F1 score: e.g., PolyLaneNet is 2nd-worst based on 3D-L1/L2 distance metrics in the attack scenarios, but in E2E-LD it is the best. With our PSLD, such judgment is strictly consistent with E2E-LD (Table~\ref{tbl:psld_summary}). One reason we observe is that the 3D-L1/L2 metrics can be greatly biased by farther points; those points by design have much less impact on the downstream AD control, but suffer from more detection errors (due to the far distance). One thought is to assign smaller weights to farther points, but how to systematically decide such weights without any form of control simulation is fundamentally nontrivial. Additionally, such a weight-based design can still be fundamentally limited in achieving sufficient AD control relating capabilities.
}

\begin{table}[h!]
\footnotesize
\vspace{-0.05in}
\centering
\setlength{\tabcolsep}{3pt}
\caption{Pearson correlation coefficient $r$ with E2E-LD. \textit{3D-L1/L2} denote the L1/L2 distances in 3D space following Reviewer 1's suggestion. \textbf{Bold} and \underline{underline} denote highest and lowest scores.
}
\vspace{-0.13in}
\begin{tabular}{lccclccc}
\toprule
            & \multicolumn{3}{c}{Benign} &  & \multicolumn{3}{c}{Attack} \\ \cline{2-4} \cline{6-8} 
            & PSLD (\textit{ours})  & 3D-L1   & 3D-L2   &  & PSLD & 3D-L1   & 3D-L2   \\ \cline{1-4} \cline{6-8} 
SCNN        & \textbf{0.93} & 0.71 & \underline{0.65} &  & \textbf{0.96} & 0.38 & \underline{0.34} \\
UltraFast   & \textbf{0.54} & 0.24 & \underline{0.19} &  & \textbf{0.93} & 0.24 & \underline{0.21} \\
PolyLaneNet & \textbf{0.49} & 0.47 & \underline{0.44} &  & \textbf{0.97} & 0.33 & \underline{0.38} \\
LaneATT     & \textbf{0.38} & 0.23 & \underline{0.17} &  & \textbf{0.95} & \underline{0.23} & \underline{0.23} \\ \hline
Average        & \textbf{0.59} & 0.41 & \underline{0.36} & & \textbf{0.95}  & \underline{0.29} & \underline{0.29}\\
\toprule
\end{tabular}
\label{tbl:3D-L1L2}
\vspace{-0.12in}
\end{table}

\section{Details of OpenPilot ALC and its integration with lane detection models} \label{appendix:openpilot}

In this section, we explain the details of OpenPilot ALC~\cite{openpilot} and the details of its integration with the 4 lane detection models we evaluate in this study. As described in~\cite{sato2020hold}, the OpenPilot ALC system consists of 3 steps: lane detection, lateral control, and vehicle actuation.

\subsection{Lane detection} 

The image frame from the front camera is input to the lane detection model in every frame (20Hz).  Since the original OpenPilot lane detection model is a recurrent neural network model, the recurrent input from the previous frame is fed to the model with the image.  All 4 models used in this study do not have a recurrent structure, i.e., they detect lanes only in the current frame.  This is because the TuSimple Challenge has a runtime limit of less than 200 ms for each frame. Another famous dataset, CULane~\cite{pan2018spatial}, does not provide even continuous frames. In autonomous driving, the recurrent structure is a reasonable choice since past frame information is always available. Hence, the run-time calculation latency imposed in the TuSimple challenge is one of the gaps between the practicality for autonomous driving and the conventional evaluation.

\subsection{Lateral control} 

Based on the detected lane line, the lateral control decides steering angle decisions to follow the lane center (i.e., the desired driving path or waypoints) as much as possible. The original OpenPilot model outputs 3 line information: left lane line, right lane line, and driving path. The desired driving path is calculated as the average of the driving path and the center line of the left and right lane lines. The steering decision is decided by the model predictive control (MPC)~\cite{MPC}. The detected lane lines are represented in the bird's-eye-view (BEV) because the steering decision needs to be decided in a world coordinate system. 

On the contrary, all 4 models used in this study detect the lane lines in the front-camera view. We thus project the detected lane lines into the BEV space with perspective transformation~\cite{hartley2003perspective, tanaka2011perspective}. The transformation matrix for this projection is created manually based on road objects such as lane markings, and then calibrated to be able to drive in a straight lane. We create the transformation matrix for each scenario as the position of the camera and the tilt of the ground are different for each scenario. The desired driving path is calculated by the average of the left and right lane lines and fed to the MPC to decide the steering angle decisions.

In addition to the desired driving path, the MPC receives the current speed and steering angle to decide the steering angle decisions. For the steering angle, we use the human driver's steering angle in the Comma2k19 dataset in the first frame. In the following frames, the steering angle is updated by the kinematic bicycle model~\cite{bicyclemodel}, which is the most widely-used motion model for vehicles. For the vehicle speed, we use the speed of human driving in the in the comma2k19 dataset as we assume that the vehicle speed is not changed largely in the free-flow scenario, in which a vehicle has at least 5--9 seconds clear headway~\cite{boora2017identification}.

\subsection{Vehicle actuation}

The step sends steering change messages to the vehicle based on the steering angle decisions. In OpenPlot, this step operates at 100 Hz control frequency. As the lane detection and lateral control outputs the steering angle decisions in 20 Hz, the vehicle actuation sends 5 messages every steering angle decision. The  steering changes are limited to a maximum value due to the physical constraints of vehicle and for stable and for stability and safety. In this study, we limit the steering angle change to 0.25$^\circ$ following prior work, which is the steering limit for production ALC systems~\cite{sato2020hold}.

We update the vehicle states with the kinematic bicycle model based on the steering change. Note that like all motion models, the kinematic bicycle model does have approximation errors to the real vehicle dynamics~\cite{kong2015kinematic}. However, more accurate motion models require more complex parameters such as vehicle mass, tire stiffness, and air resistance~\cite{mathwork_motionmodel}. In this study, since our focus is on understanding the impact of lane detection model robustness on end-to-end driving, the most widely-used kinematic bicycle model is a sufficient choice for simulating closed-loop control behaviors.

\section{Additional Discussions and Results}
\label{appendx:more_end2end}

\subsection{Additional Discussions}

\newpart{
\textbf{Ground-Truth Road Center.} 
We obtain the ground truth waypoints based on the human driving traces. Ideally, the waypoints should be obtained by measuring roads. However, since this study focuses on the general trends of the 4 lane detection approaches, we consider that the impact of this factor should not have a major effect. If you want to use PSLD to capture more subtle differences between models, the ground truth should be more accurate.

\textbf{Differentiable PSLD Regularization.} 
We show that PSLD works as a good surrogate for E2E-LD. Next, we may want to minimize this metric directly in the model training. Since the only non-differentiable computation in PSLD is the lateral controller, we can replace this part with a differentiable controller~\cite{tassa2014control, amos2018differentiable} and incorporate it as a regularization term in the loss function for training. Detailed study of this problem is left to future work.
}

\subsection{TuSimple Challenge Dataset}
\begin{figure}[h!]
\centering
\includegraphics[width=\linewidth]{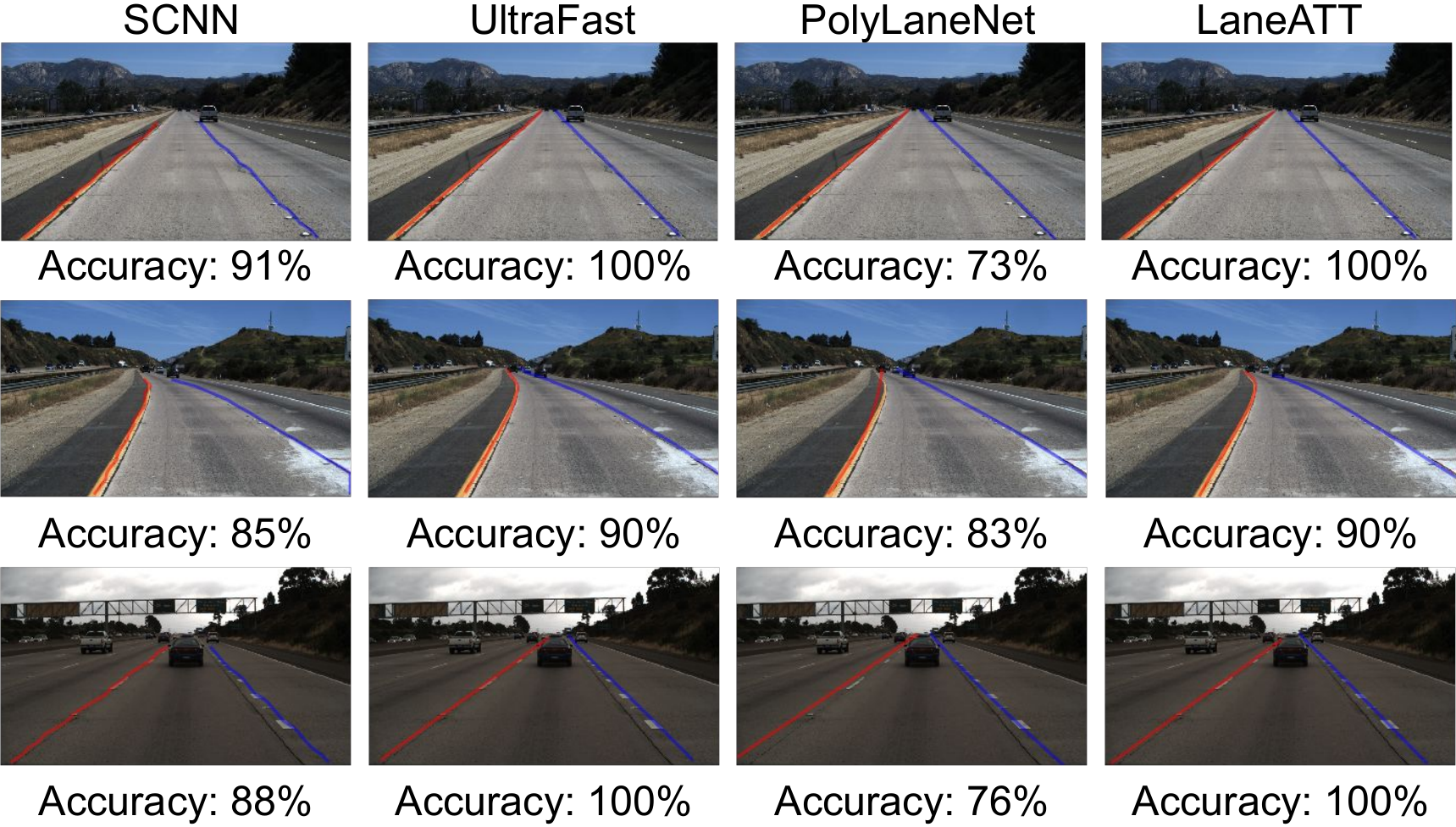}
\vspace{-0.2in}
\caption{\newpart{Examples of lane detection results and the accuracy metric in benign scenarios on TuSimple Challenge dataset~\cite{tusimple}. As shown, the conventional accuracy metric does not necessarily indicate drivability if used in autonomous driving}}
\label{fig:benign_tusimple}
\end{figure}

\newpart{
Fig.~\ref{fig:benign_tusimple} shows the examples of lane detection results and the accuracy metric in benign scenarios on the TuSimple Challenge dataset~\cite{tusimple}. The limitations of the conventional metrics can be found in benign cases as well. As shown, SCNN has always higher accuracy than PolyLaneNet (at most 18\% edge). Such a large leading edge is across the dataset as in Table~\ref{tbl:tusimple} (89\% vs 72\% in Accuracy, 75\% vs 50\% in F1 score). However, for downstream AD their performances are almost the same, with PolyLaneNet actually slightly better (Table~\ref{tbl:summary_old_metric} of the main paper).
}

\subsection{Comma2k19 LD Dataset}
\begin{figure}[htb]
\begin{center}
    \includegraphics[width=\linewidth]{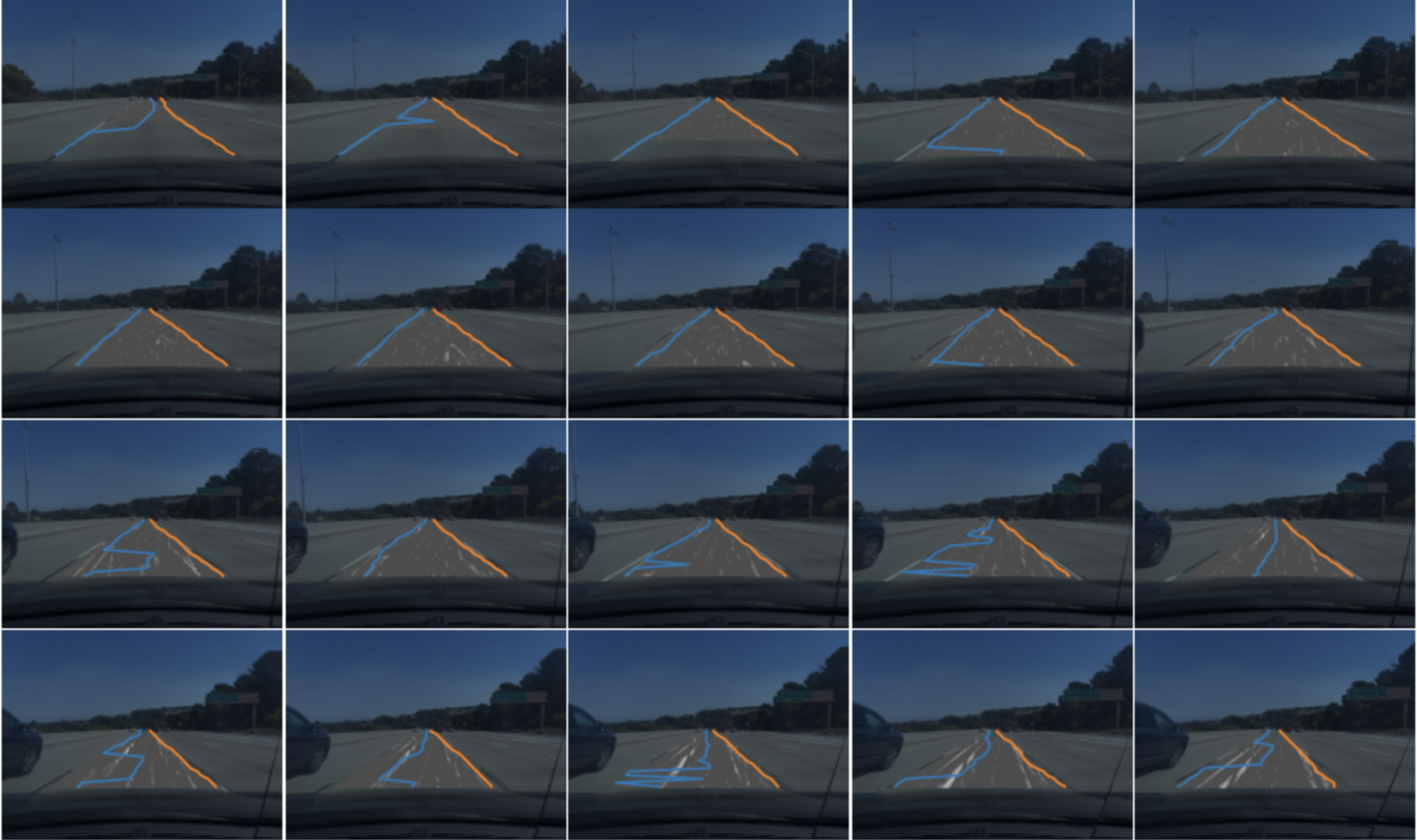}
\end{center}
\vspace{-0.2in}
\caption{
The first 20 frames (from left-top to right-bottom) of an attack scenario on \textbf{SCNN}. The vehicle is deviating to right due to the attack.
}
\label{fig:demo_scnn}
\end{figure}

\begin{figure}[htb]
\begin{center}
    \includegraphics[width=\linewidth]{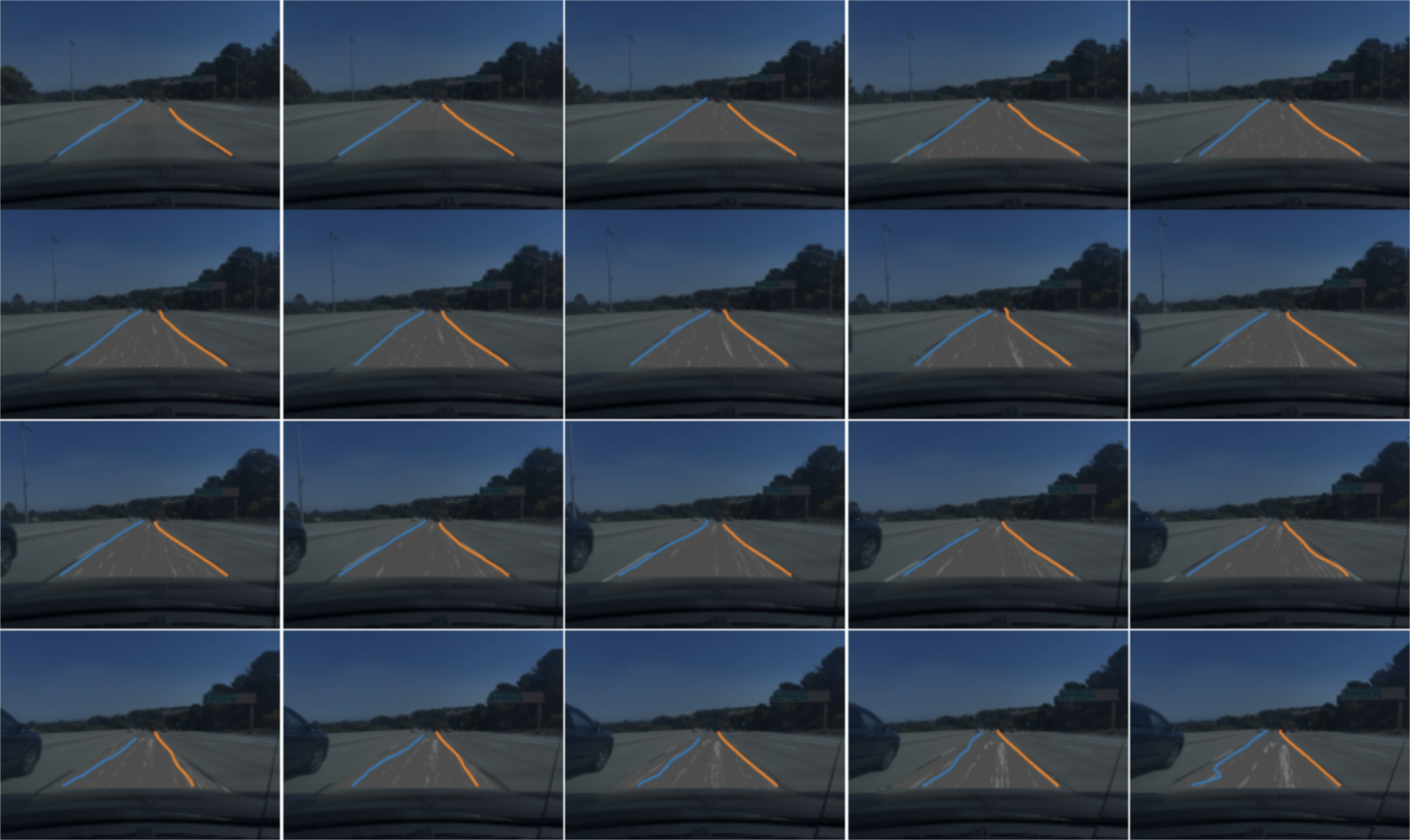}
\end{center}
\vspace{-0.2in}
\caption{
The first 20 frames (from left-top to right-bottom) of an attack scenario on \textbf{UltraFast}. The vehicle is deviating to right due to the attack.
}
\label{fig:demo_ultrafast}
\end{figure}

\begin{figure}[htb]
\begin{center}
    \includegraphics[width=\linewidth]{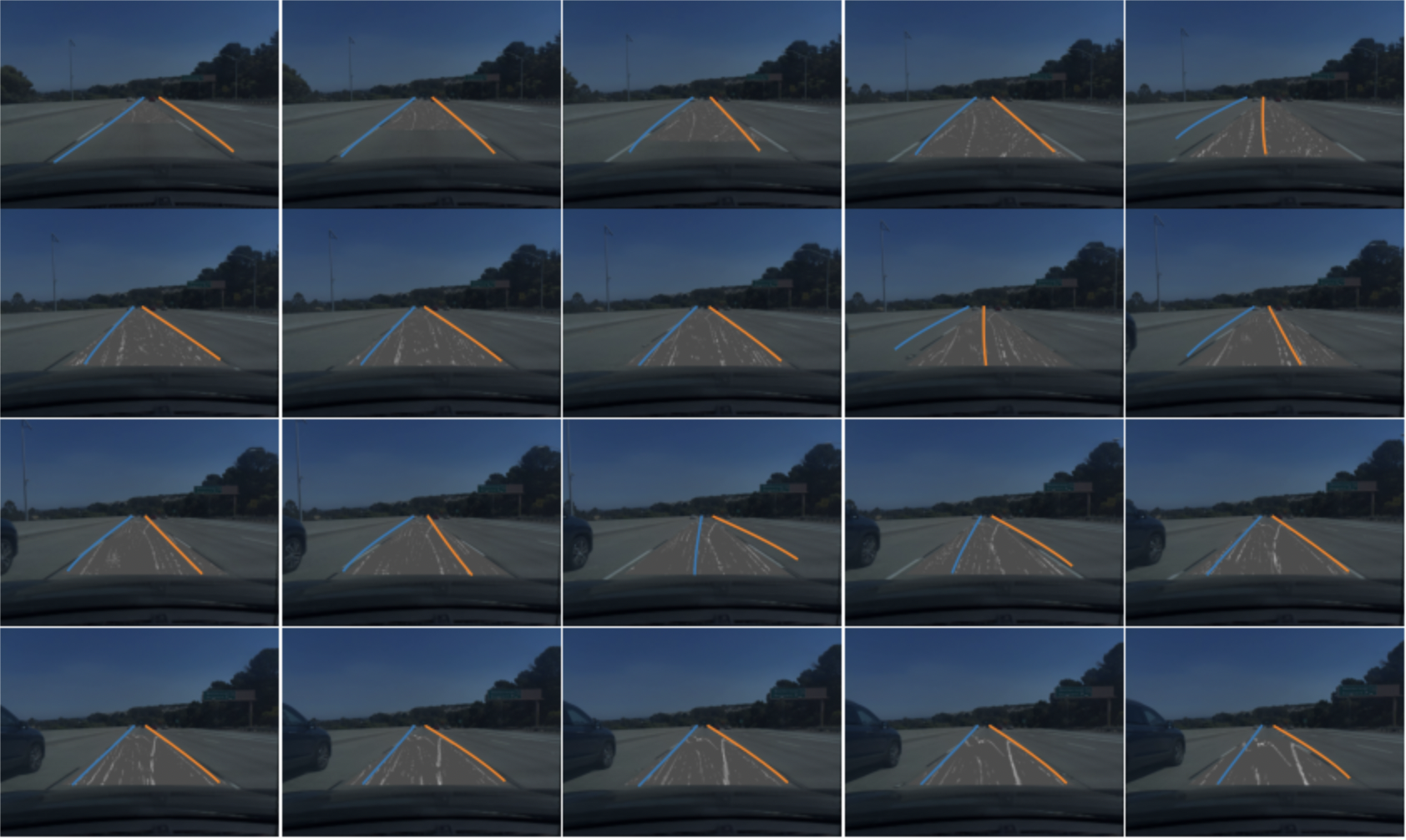}
\end{center}
\vspace{-0.2in}
\caption{
The first 20 frames (from left-top to right-bottom) of an attack scenario on \textbf{PolyLaneNet}. The vehicle is deviating to right due to the attack.
}
\label{fig:demo_polylanenet}
\vspace{-0.1in}
\end{figure}

\begin{figure}[htb]
\begin{center}
    \includegraphics[width=\linewidth]{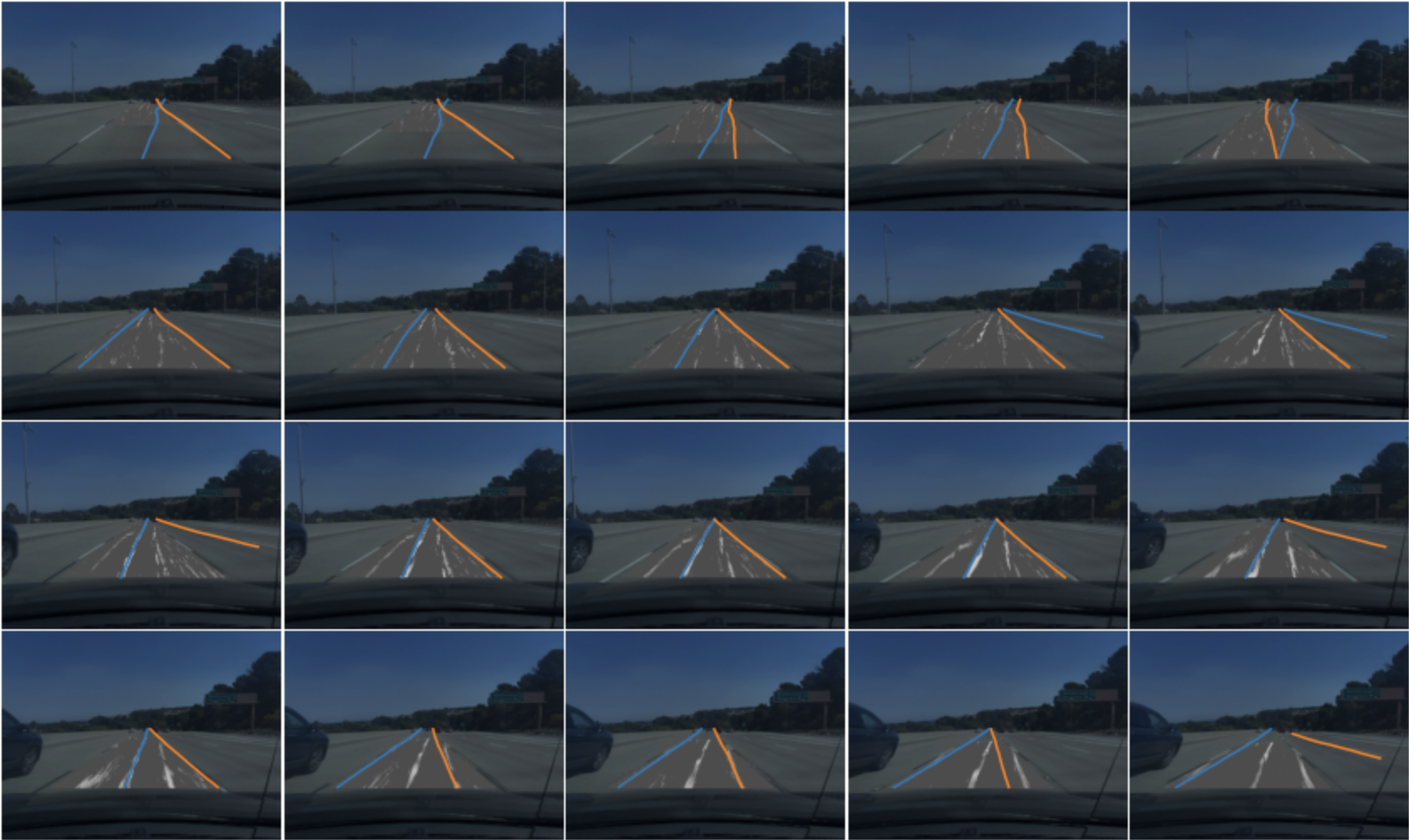}
\end{center}
\vspace{-0.2in}
\caption{
The first 20 frames (from left-top to right-bottom) of an attack scenario on \textbf{LaneATT}. The vehicle is deviating to right due to the attack.
}
\label{fig:demo_laneatt}
\vspace{-0.1in}
\end{figure}

We synthesize font-camera frames with a vehicle motion model~\cite{bicyclemodel} and perspective transformation~\cite{hartley2003perspective, tanaka2011perspective}. Fig.~\ref{fig:demo_scnn},~\ref{fig:demo_ultrafast},~\ref{fig:demo_polylanenet},~\ref{fig:demo_laneatt}, 
show the first 20 frames under attack and their detection results of the 4 lane detection methods, respectively. As shown, the generated images are generally complete and the distortion is very slight.

%% file: main.bbl
\begin{thebibliography}{10}\itemsep=-1pt

\bibitem{apollo}
{Baidu Apollo}.
\newblock \url{https://github.com/ApolloAuto/apollo}.

\bibitem{coursera}
{Introduction to Self-Driving Cars}.
\newblock \url{https://www.coursera.org/learn/intro-self-driving-cars}.

\bibitem{lgsvl}
{LGSVL Simulator: An Autonomous Vehicle Simulator}.
\newblock \url{https://github.com/lgsvl/simulator/}.

\bibitem{mathwork_motionmodel}
{Modeling a Vehicle Dynamics System}.
\newblock
  \url{https://www.mathworks.com/help/ident/ug/modeling-a-vehicle-dynamics-system.html}.

\bibitem{openpilot}
{OpenPilot: Open Source Driving Agent}.
\newblock \url{https://github.com/commaai/openpilot}.

\bibitem{gmsupercruise}
{Super Cruise - Hands Free Driving | Cadillac Ownership}.
\newblock
  \url{https://www.cadillac.com/world-of-cadillac/innovation/super-cruise}.

\bibitem{tesla2020autopilot}
{Tesla Autopilot}.
\newblock \url{https://www.tesla.com/autopilot}.

\bibitem{sae2018}
{Taxonomy and Definitions for Terms Related to Driving Automation Systems for
  On-Road Motor Vehicles}.
\newblock {\em SAE International,(J3016)}, 2016.

\bibitem{tusimple}
{TuSimple Lane Detection Challenge}.
\newblock
  \url{https://github.com/TuSimple/tusimple-benchmark/tree/master/doc/lane_detection},
  2017.

\bibitem{waymo_open_dataset}
{Waymo Open Dataset: An Autonomous Driving Dataset}.
\newblock \url{https://www.waymo.com/open}, 2019.

\bibitem{simlink2020lane}
{Lane Keeping Assist System Using Model Predictive Control}.
\newblock
  \url{https://www.mathworks.com/help/mpc/ug/lane-keeping-assist-system-using-model-predictive-control.html},
  2020.

\bibitem{Comma2k19-LD}
{Comma2k19 LD}.
\newblock \url{https://www.kaggle.com/tkm2261/comma2k19-ld}, 2022.

\bibitem{amos2018differentiable}
Brandon Amos, Ivan Jimenez, Jacob Sacks, Byron Boots, and J~Zico Kolter.
\newblock {Differentiable MPC for end-to-end planning and control}.
\newblock In {\em NeurIPS}, 2018.

\bibitem{bansal2019chauffeurnet}
Mayank Bansal, Alex Krizhevsky, and Abhijit~S. Ogale.
\newblock {ChauffeurNet: Learning to Drive by Imitating the Best and
  Synthesizing the Worst}.
\newblock In {\em RSS}, 2019.

\bibitem{llamas2019}
Karsten Behrendt and Ryan Soussan.
\newblock {Unsupervised Labeled Lane Marker Dataset Generation Using Maps}.
\newblock In {\em IEEE International Conference on Computer Vision}, 2019.

\bibitem{bojarski2016end}
Mariusz Bojarski, Davide Del~Testa, Daniel Dworakowski, Bernhard Firner, Beat
  Flepp, Prasoon Goyal, Lawrence~D Jackel, Mathew Monfort, Urs Muller, Jiakai
  Zhang, et~al.
\newblock {End to End Learning for Self-Driving Cars}.
\newblock {\em arXiv preprint arXiv:1604.07316}, 2016.

\bibitem{boora2017identification}
Amardeep Boora, Indrajit Ghosh, and Satish Chandra.
\newblock Identification of free flowing vehicles on two lane intercity
  highways under heterogeneous traffic condition.
\newblock {\em Transportation Research Procedia}, 21:130--140, 2017.

\bibitem{brown2017advpatch}
Tom Brown, Dandelion Mane, Aurko Roy, Martin Abadi, and Justin Gilmer.
\newblock {Adversarial Patch}.
\newblock {\em arXiv preprint arXiv:1712.09665}, 2017.

\bibitem{casas2021mp3}
Sergio Casas, Abbas Sadat, and Raquel Urtasun.
\newblock {MP3: A Unified Model to Map, Perceive, Predict and Plan}.
\newblock In {\em CVPR}, 2021.

\bibitem{chang2019argoverse}
Ming-Fang Chang, John Lambert, Patsorn Sangkloy, Jagjeet Singh, Slawomir Bak,
  Andrew Hartnett, De Wang, Peter Carr, Simon Lucey, Deva Ramanan, et~al.
\newblock {Argoverse: 3D Tracking and Forecasting with Rich Maps}.
\newblock In {\em CVPR}, 2019.

\bibitem{chen2019med3d}
Sihong Chen, Kai Ma, and Yefeng Zheng.
\newblock {Med3d: Transfer Learning for 3D Medical Image Analysis}.
\newblock {\em arXiv preprint arXiv:1904.00625}, 2019.

\bibitem{bert}
Jacob Devlin, Ming-Wei Chang, Kenton Lee, and Kristina Toutanova.
\newblock {{BERT}: Pre-training of Deep Bidirectional Transformers for Language
  Understanding}.
\newblock In {\em NAACL}, 2019.

\bibitem{dorf2011modern}
Richard~C Dorf and Robert~H Bishop.
\newblock {\em {Modern Control Systems}}.
\newblock Pearson, 2011.

\bibitem{dosovitskiy2017carla}
Alexey Dosovitskiy, German Ros, Felipe Codevilla, Antonio Lopez, and Vladlen
  Koltun.
\newblock {{CARLA}: {An} Open Urban Driving Simulator}.
\newblock In {\em CoRL}, 2017.

\bibitem{carla}
Alexey Dosovitskiy, German Ros, Felipe Codevilla, Antonio Lopez, and Vladlen
  Koltun.
\newblock {CARLA: An Open Urban Driving Simulator}.
\newblock In {\em CoRL}, pages 1--16, 2017.

\bibitem{eykholt2018physical}
Kevin Eykholt, Ivan Evtimov, Earlence Fernandes, Bo Li, Amir Rahmati, Florian
  Tramer, Atul Prakash, Tadayoshi Kohno, and Dawn Song.
\newblock {Physical Adversarial Examples for Object Detectors}.
\newblock In {\em WOOT}, 2018.

\bibitem{goodfellow2014explaining}
Ian~J Goodfellow, Jonathon Shlens, and Christian Szegedy.
\newblock {Explaining and Harnessing Adversarial Examples}.
\newblock {\em arXiv preprint arXiv:1412.6572}, 2014.

\bibitem{hartley2003perspective}
Richard Hartley and Andrew Zisserman.
\newblock {\em {Multiple View Geometry in Computer Vision}}.
\newblock Cambridge University Press, 2 edition, 2003.

\bibitem{hillel2014recent}
Aharon~Bar Hillel, Ronen Lerner, Dan Levi, and Guy Raz.
\newblock {Recent Progress in Road and Lane Detection: A Survey}.
\newblock {\em Machine vision and applications}, 25(3):727--745, 2014.

\bibitem{hou2020inter}
Yuenan Hou, Zheng Ma, Chunxiao Liu, Tak-Wai Hui, and Chen~Change Loy.
\newblock {Inter-Region Affinity Distillation for Road Marking Segmentation}.
\newblock In {\em CVPR}, 2020.

\bibitem{hou2019learning}
Yuenan Hou, Zheng Ma, Chunxiao Liu, and Chen~Change Loy.
\newblock {Learning Lightweight Lane Detection CNNs by Self Attention
  Distillation}.
\newblock In {\em CVPR}, 2019.

\bibitem{hsu2018learning}
Yen-Chang Hsu, Zheng Xu, Zsolt Kira, and Jiawei Huang.
\newblock {Learning to Cluster for Proposal-Free Instance Segmentation}.
\newblock In {\em IJCNN}, 2018.

\bibitem{jain2021autonomy}
Ashesh Jain, Luca Del~Pero, Hugo Grimmett, and Peter Ondruska.
\newblock {Autonomy 2.0: Why Is Self-Driving Always 5 Years Away?}
\newblock {\em arXiv preprint arXiv:2107.08142}, 2021.

\bibitem{jing2021tencent}
Pengfei Jing, Qiyi Tang, Yuefeng Du, Lei Xue, Xiapu Luo, Ting Wang, Sen Nie,
  and Shi Wu.
\newblock {Too Good to Be Safe: Tricking Lane Detection in Autonomous Driving
  with Crafted Perturbations}.
\newblock In {\em USENIX Security}, 2021.

\bibitem{kong2015kinematic}
Jason Kong, Mark Pfeiffer, Georg Schildbach, and Francesco Borrelli.
\newblock {Kinematic and Dynamic Vehicle Models for Autonomous Driving Control
  Design}.
\newblock In {\em IV}, 2015.

\bibitem{lee2012unified}
Jin-Woo Lee and Bakhtiar Litkouhi.
\newblock {A Unified Framework of the Automated Lane Centering/Changing Control
  for Motion Smoothness Adaptation}.
\newblock In {\em ITSC}, 2012.

\bibitem{li2019line}
Xiang Li, Jun Li, Xiaolin Hu, and Jian Yang.
\newblock {Line-CNN: End-to-End Traffic Line Detection with Line Proposal
  Unit}.
\newblock {\em ITSC}, 2019.

\bibitem{liu2021CondLaneNet}
Lizhe Liu, Xiaohao Chen, Siyu Zhu, and Ping Tan.
\newblock {CondLaneNet: A Top-To-Down Lane Detection Framework Based on
  Conditional Convolution}.
\newblock In {\em ICCV}, 2021.

\bibitem{neven2018towards}
Davy Neven, Bert De~Brabandere, Stamatios Georgoulis, Marc Proesmans, and Luc
  Van~Gool.
\newblock {Towards End-to-End Lane Detection: An Instance Segmentation
  Approach}.
\newblock In {\em IV}, 2018.

\bibitem{pan2018spatial}
Xingang Pan, Jianping Shi, Ping Luo, Xiaogang Wang, and Xiaoou Tang.
\newblock {Spatial as Deep: Spatial CNN for Traffic Scene Understanding}.
\newblock In {\em AAAI}, 2018.

\bibitem{philion2019fastdraw}
Jonah Philion.
\newblock {FastDraw: Addressing the Long Tail of Lane Detection by Adapting a
  Sequential Prediction Network}.
\newblock In {\em CVPR}, 2019.

\bibitem{qin2020ultra}
{Qin, Zequn and Wang, Huanyu and Li, Xi}.
\newblock {Ultra Fast Structure-Aware Deep Lane Detection}.
\newblock In {\em ECCV}, 2020.

\bibitem{qu2021fololane}
Zhan Qu, Huan Jin, Yang Zhou, Zhen Yang, and Wei Zhang.
\newblock {Focus on Local: Detecting Lane Marker From Bottom Up via Key Point}.
\newblock In {\em CVPR}, 2021.

\bibitem{bicyclemodel}
Rajesh Rajamani.
\newblock {\em Vehicle Dynamics and Control}.
\newblock Springer Science \& Business Media, 2011.

\bibitem{ren2015faster}
Shaoqing Ren, Kaiming He, Ross Girshick, and Jian Sun.
\newblock {Faster R-CNN: Towards Real-Time Object Detection with Region
  Proposal Networks}.
\newblock In {\em NeurIPS}, 2015.

\bibitem{MPC}
{Richalet, J. and Rault, A. and Testud, J. L. and Papon, J.}
\newblock {Paper: Model Predictive Heuristic Control}.
\newblock {\em Automatica}, 14(5):413–428, Sept. 1978.

\bibitem{sato2020hold}
Takami Sato, Junjie Shen, Ningfei Wang, Yunhan Jia, Xue Lin, and Qi~Alfred
  Chen.
\newblock {Dirty Road Can Attack: Security of Deep Learning based Automated
  Lane Centering under Physical-World Attack}.
\newblock {\em USENIX Security Symposium}, 2021.

\bibitem{comma2k19}
Harald Schafer, Eder Santana, Andrew Haden, and Riccardo Biasini.
\newblock {A Commute in Data: The comma2k19 Dataset}.
\newblock {\em arXiv preprint arXiv:1812.05752}, 2018.

\bibitem{Szegedy2014}
Christian Szegedy, Wojciech Zaremba, Ilya Sutskever, Joan Bruna, Dumitru Erhan,
  Ian Goodfellow, and Rob Fergus.
\newblock {Intriguing Properties of Neural Networks}.
\newblock In {\em ICLR}, 2014.

\bibitem{tabelini2021cvpr}
Lucas Tabelini, Rodrigo Berriel, Thiago M.~Paix\ ao, Claudine Badue, Alberto
  Ferreira~De Souza, and Thiago Oliveira-Santos.
\newblock {Keep your Eyes on the Lane: Real-time Attention-guided Lane
  Detection}.
\newblock In {\em CVPR}, 2021.

\bibitem{tabelini2021polylanenet}
Lucas Tabelini, Rodrigo Berriel, Thiago~M Paixao, Claudine Badue, Alberto~F
  De~Souza, and Thiago Oliveira-Santos.
\newblock {Polylanenet: Lane Estimation via Deep Polynomial Regression}.
\newblock In {\em ICPR}, 2021.

\bibitem{tanaka2011perspective}
Shiho Tanaka, Kenichi Yamada, Toshio Ito, and Takenao Ohkawa.
\newblock {Vehicle Detection Based on Perspective Transformation Using
  Rear-View Camera}.
\newblock {\em Hindawi Publishing Corporation International Journal of
  Vehicular Technology}, 9, 03 2011.

\bibitem{TANG2021107623}
Jigang Tang, Songbin Li, and Peng Liu.
\newblock {A Review of Lane Detection Methods Based on Deep Learning}.
\newblock {\em Pattern Recognition}, 111:107623, 2021.

\bibitem{tassa2014control}
Yuval Tassa, Nicolas Mansard, and Emo Todorov.
\newblock Control-limited differential dynamic programming.
\newblock In {\em ICRA}, 2014.

\bibitem{watzenig2016automated}
Daniel Watzenig and Martin Horn.
\newblock {\em {Automated Driving: Safer and More Efficient Future Driving}}.
\newblock Springer, 2016.

\bibitem{yoo2020end}
Seungwoo Yoo, Hee~Seok Lee, Heesoo Myeong, Sungrack Yun, Hyoungwoo Park,
  Janghoon Cho, and Duck~Hoon Kim.
\newblock {End-to-End Lane Marker Detection via Row-Wise Classification}.
\newblock In {\em CVPR Workshops}, 2020.

\bibitem{yu2020bdd100k}
Fisher Yu, Haofeng Chen, Xin Wang, Wenqi Xian, Yingying Chen, Fangchen Liu,
  Vashisht Madhavan, and Trevor Darrell.
\newblock {Bdd100k: A Diverse Driving Dataset for Heterogeneous Multitask
  Learning}.
\newblock In {\em CVPR}, 2020.

\bibitem{zheng2020resa}
Tu Zheng, Hao Fang, Yi Zhang, Wenjian Tang, Zheng Yang, Haifeng Liu, and Deng
  Cai.
\newblock {RESA: Recurrent Feature-Shift Aggregator for Lane Detection}, 2020.

\bibitem{zheng2021RESA}
Tu Zheng, Hao Fang, Yi Zhang, Wenjian Tang, Zheng Yang, Haifeng Liu, and Deng
  Cai.
\newblock {RESA: Recurrent Feature-Shift Aggregator for Lane Detection}.
\newblock {\em AAAI}, 2021.

\end{thebibliography}
